%% file: arxiv_svm.tex
\documentclass[10pt,twocolumn,letterpaper]{article}

\usepackage[pagenumbers]{cvpr} 

\input{preamble}

\definecolor{cvprblue}{rgb}{0.21,0.49,0.74}
\usepackage[pagebackref,breaklinks,colorlinks,citecolor=cvprblue]{hyperref}

\def\paperID{1143} 
\def\confName{CVPR}
\def\confYear{2024}

\title{Small, Versatile and Mighty: A Range-View Perception Framework}

\author{Qiang Meng\textsuperscript{1} \quad 
        Xiao Wang\textsuperscript{1} \quad
        JiaBao Wang\textsuperscript{2} \quad
        Liujiang Yan\textsuperscript{1} \quad  
        Ke Wang\textsuperscript{1} \quad 
        \\
        \textsuperscript{1} KargoBot \quad
        \textsuperscript{2} VCIP, School of Computer Science, Nankai University        
        \\
        {\tt\small
        irvingmeng@outlook.com
        } 
}

\begin{document}
\maketitle

\input{0_abstract}    
\input{1_introduction}

\input{2_related}
\input{3_methods}
\input{4_experiments}

\input{5_conclusions}
\clearpage
\input{6_supplementary}

{
    \small
    \bibliographystyle{ieeenat_fullname}
    \bibliography{egbib}
}


\end{document}

%% file: preamble.tex
\usepackage[dvipsnames]{xcolor}

\usepackage{multirow}
\usepackage{multicol}
\usepackage{bm}

\usepackage[
skip=5pt]{caption}
\usepackage[ruled,vlined,linesnumbered]{algorithm2e}
\SetKwInOut{KwIn}{Input}
\SetKwInOut{KwOut}{Output}
\SetKwInOut{KwPara}{Parameters}

%% file: 0_abstract.tex
\begin{abstract}

  Despite its compactness and information integrity, the range view representation of LiDAR data rarely occurs as the first choice for 3D perception tasks.
  In this work, we further push the envelop of the range-view representation with a novel multi-task framework, achieving unprecedented 3D detection performances.
  Our proposed \textit{Small, Versatile, and Mighty} (SVM) network utilizes a pure convolutional architecture to fully unleash the efficiency and multi-tasking potentials of the range view representation.
  To boost detection performances, we first propose a range-view specific Perspective Centric Label Assignment (PCLA) strategy, and a novel View Adaptive Regression (VAR) module to further refine hard-to-predict box properties.
  In addition, our framework seamlessly integrates semantic segmentation and panoptic segmentation tasks for the LiDAR point cloud, without extra modules.
  Among range-view-based methods, our model achieves new state-of-the-art detection performances on the Waymo Open Dataset. 
  Especially, over 10 mAP improvement over convolutional counterparts can be obtained on the vehicle class.
  Our presented results for other tasks further reveal the multi-task capabilities of the proposed small but mighty framework.

\end{abstract}
    

%% file: 1_introduction.tex
\section{Introduction}\label{sec:intro}

\begin{figure}[htb!]
    \centering
    \includegraphics[trim={80pt 75pt 200pt 90pt},clip, width=0.48\textwidth]{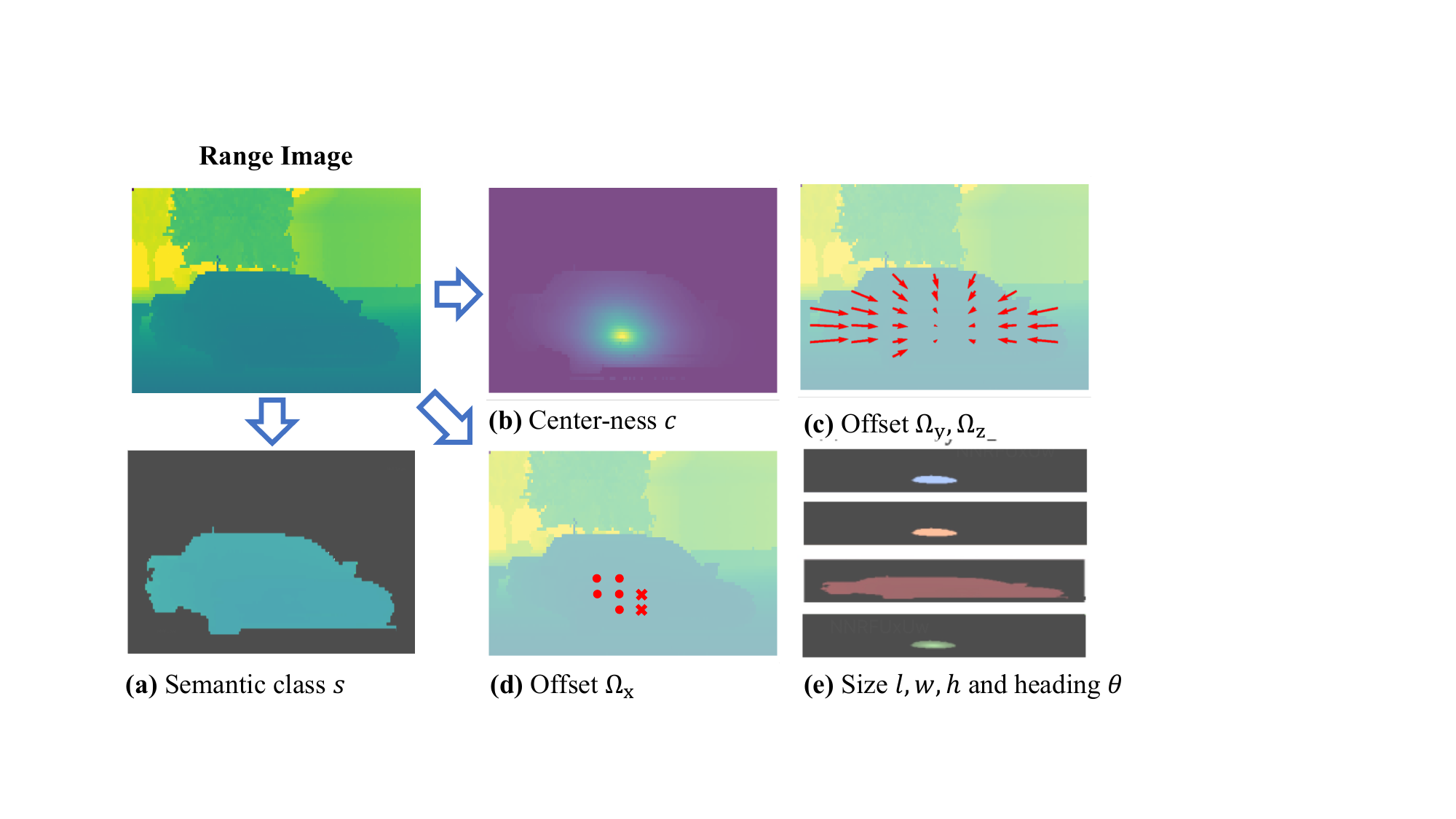}
    \caption{
      Our range-view-based framework generates predictions including:
      Semantic class $s$ for each valid point, facilitating the task of semantic segmentation;
      Center-ness $c$, offsets $\Omega_y$, $\Omega_z$, and box height $h$ for each foreground point, with the potential to perform panoptic segmentation;
      Remaining elements for 3D object detection are regressed for centric points within boxes in the perspective view.
      Here, $\bm{\scriptstyle \times}$ and $\scriptstyle\bullet$ in (d) represent directions inward and outward perpendicular to the plane, respectively.
    }
    \label{fig:intro}
  \end{figure}

  The LiDAR range sensor plays a vital role in safety-critical applications (\eg, object detection and panoptic segmentation in autonomous driving) by providing accurate 3D environment measurements independent of lighting conditions.
  Nonetheless, the LiDAR point cloud is inherently non-uniform, orderless, and sparse, prohibiting direct applications of highly-optimized operators such as convolutions.
  One way to address this issue is to first establish a neighborhood structure in the point clouds, via expensive radius search or nearest neighbor search, then performant convolution operators can be applied in the local neighborhood~\cite{chen2019fast,yang2018ipod,shi2020point,qi2019deep}.
  Another approach is to create regular voxel grids~\cite{zhou2018voxelnet, yan2018second, deng2021voxel, zhou2022centerformer, zhang2020polarnet} or pillars of voxels~\cite{yin2021center, lang2019pointpillars, li2023pillarnext, shi2019part, zhou2022centerformer} from the input points through quantization, with inevitable information loss.
  Despite their great success, algorithms utilizing point sets and voxel grids commonly necessitate heavy computations, posing challenges in scaling them for real-time autonomous systems.
  In contrast, the range image organizes 3D data into a structured 2D visual representation in a lossless fashion.
  As a result, the range image is unarguably the most compact and efficient one among all data representations of LiDAR point clouds.

  In addition to the efficiency advantage, another underestimated benefit of the range-view representation lies in its potential for multiple tasks.
  Compared to the dominant grid representation, range-view has significantly higher coherency between segmentation and detection tasks.
  We believe the following aspects attribute to the reduced discrepancy:
  (1) Resolution requirement:
  Current voxel-grid-based multi-task frameworks~\cite{xu2023aop,zhou2023lidarformer,ye2023lidarmultinet} predict separate heatmaps for different tasks.
  Notably, the heatmap resolution for segmentation is substantially higher than that required for object detection.
  On the contrary, range-view-based detectors~\cite{sun2021rsn, fan2021rangedet, meyer2019lasernet, tian2022fully} usually produce heatmaps with similar resolution as the input range image, inherently meeting the resolution requirement for segmentation.
  (2) Foreground definition: 
  Most range-view-based detectors~\cite{sun2021rsn, fan2021rangedet, meyer2019lasernet, tian2022fully} consider all points within bounding boxes as their foregrounds. 
  Unlike camera images, foregrounds defined by points within boxes can exclude background pixels in point clouds because of the known 3D coordinates of all points and boxes.
  This property naturally leads to foreground points forming the semantic labels, as illustrated in \cref{fig:intro}(a).
  (3) Representation density:
  LiDAR point cloud is inherently sparse, and the quantized voxel grid contains large portions of empty cells.
  For dense prediction tasks like segmentation, such emptiness in the input representation and the corresponding feature maps lead to reduced receptive fields, making the inference of high-level semantics much harder.
 
  This work aims to highlight the advantages of the range-view representation by proposing a novel multi-task framework with a simple fully convolutional architecture.
  Previous works mainly improve detection performances by developing customized kernels~\cite{fan2021rangedet,chai2021point}, using multiple and dedicated detection heads~\cite{fan2021rangedet,tian2022fully}, or running complex post-processing~\cite{meyer2019lasernet}.
  While these methods achieve performance improvements, they sacrifice the efficiency of the range-view representation.
  In contrast, we aim to improve task performance on simple architectures using insightful module designs and careful training strategies.
  For classification, we propose the module of Perspective Centric Label Assignment (PCLA) to predict semantic classes and perspective center-nesses, as shown in \cref{fig:intro}(a) and (b). 
  The semantic classes can contribute to segmentation tasks as mentioned earlier, while the center-nesses will benefit object detection by filtering erroneous predictions.

  For the regression, we begin with analyzing the learning difficulties of offsets $\Omega_y, \Omega_z$ in \cref{fig:intro}(c) and $\Omega_x$ in \cref{fig:intro}(d).
  We found that the former two elements are evidently easier to learn, which could be because range image compresses the dimension of the view direction while preserving the neighborhood relationships in other directions.
  This finding motivates our View Adaptive Regression (VAR) to discriminately process elements in the range view.
  Specifically, VAR groups the regression elements into those preferred by the perspective view (\ie, offsets $\Omega_y, \Omega_z$ and box height $h$) and those clearly visible in the bird's-eye-view (\ie, the offset $\Omega_x,$ box length $l$, width $w$, and yaw angle $\phi$ defined in bird's-eye-view), and adopts two branches for separate regressions.
  Additionally, VAR regresses $\{\Omega_y, \Omega_z, h\}$ for all object points as indicated in \cref{fig:intro}(c,e), which together with semantic classes and center-nesses facilitate the execution of  panoptic segmentation.
  Conversely, remaining elements are predicted for perspective-centric points, thus enhancing detection accuracy by disregarding error-prone edge points.
  
  In the end, our framework achieves the state-of-the-art detection performance among range-view-based methods. 
  Furthermore, our framework can be seamlessly applied to segmentation tasks without the need for additional modules, as validated by our experiments.
  We summarize our contributions as follows:
  \begin{enumerate}
    \itemsep0em
    \item We propose a fully convolutional and single-stage framework which highlights two vital advantages (\ie, efficiency and potential for multi-tasks) of the range-view representation.
    Without whistles and bells, the framework is capable of performing 3D object detection, semantic segmentation, and panoptic segmentation.
    \item We propose to optimize the classification and regression procedures by the PCLA and VAR modules, respectively.
    These range-view-suited schemes yield significant performance improvements in detection, surpassing convolutional counterparts by over 10 mAP. 
  \end{enumerate}



%% file: 2_related.tex
\section{Related Works}\label{sec:related_work}

\subsection{Point Cloud Representation}
LiDAR sensors collect precise and high-fidelity 3D structural information, which can be represented in diverse formats, \ie, raw points~\cite{chen2019fast,yang2018ipod,shi2020point,qi2019deep, shi2019pointrcnn, qi2018frustum, yang20203dssd, sautier2023bevcontrast,zhu2023curricular}, voxel grids~\cite{zhou2018voxelnet, yan2018second, yin2021center, lang2019pointpillars, deng2021voxel, shi2019part,fan2022embracing,dosovitskiy2020image, li2023pillarnext, zhou2022centerformer} and range images~\cite{bewley2020range, meyer2019lasernet}.
To extract features from the raw points, a neighborhood must be established for each point via radius search or nearest neighbor search. Such expensive geometric neighborhood indexing procedure prohibits the use of raw point cloud formats in real-time applications.
By quantizing the point cloud into a fixed grid of voxels~\cite{zhou2018voxelnet} or pillars of voxels~\cite{lang2019pointpillars}, regular grid representation allows for efficient processing at the cost of quantization error and information loss.
This representation has gained prominence in current literature~\cite{zhou2018voxelnet, yan2018second, yin2021center, lang2019pointpillars, deng2021voxel, shi2019part}, especially for 3D object detection.
In contrast, range image directly reflects the LiDAR sampling process, and offers a compact yet integral representation of point clouds.
The compact nature facilitates fast processing of range images, making this modality the most efficient among all data representations.

\subsection{Range-view-based Solutions}
Thanks to the similar data format between camera image and range image, many well-established concepts and techniques can be served as valuable references for range-view-based methods.
This has led to the widespread adoption of the range view in 3D semantic segmentation, with efforts focused on network architectures~\cite{behley2019semantickitti, cortinhal2020salsanext}, customized kernels~\cite{kochanov2020kprnet,xu2020squeezesegv3}, post-processing steps~\cite{milioto2019rangenet++}, \etc.
Recent rapid advancements of unsupervised learning and large vision models further propel the range-view-based semantic segmentation~\cite{kong2023rethinking, ando2023rangevit} to reach state-of-the-art performances.

While range-view-based solutions have achieved remarkable success in semantic segmentation, they still occupy a minority position within the fields of 3D object detection~\cite{meyer2019lasernet,bewley2020range,sun2021rsn,fan2021rangedet,chai2021point,tian2022fully} and panoptic segmentation~\cite{milioto2020lidar, sirohi2021efficientlps}.
This disparity can be attributed to the inherent 2.5D nature of the input range image, which mismatches the requisite for generating 3D predictions in these tasks.


\subsection{Multi-task Perception}


\begin{table}[tb!]
 \vspace*{-1em}
    \centering
    \renewcommand{\arraystretch}{1.05}
    \renewcommand{\tabcolsep}{1pt}
    \footnotesize
    \begin{tabular}{l|cccccc}
      \toprule
      \multirow{2}{*}{Method} & \multirow{2}{*}{Detection Head} & \multicolumn{2}{c}{Grid Resolution} \\
      \cline{3-4} 
        && detection & segmentation \\
        \midrule
        AOP-Net~\cite{xu2023aop} & CenterPoint~\cite{yin2021center}& $(\frac{H}{8}, \frac{W}{8}, \frac{Z}{16})$ & $(\frac{H}{2}, \frac{W}{2}, \frac{Z}{2})$ \\
        LiDARFormer~\cite{zhou2023lidarformer} & CenterFormer~\cite{zhou2022centerformer} &  $(\frac{H}{8}, \frac{W}{8})$ & \scriptsize{$(H, W, Z)$}\\
        LiDARMultiNet~\cite{ye2023lidarmultinet} &CenterPoint~\cite{yin2021center} & \scriptsize{$(H, W)$} & \scriptsize{$(H, W, Z)$}\\
      \bottomrule
    \end{tabular}
    \caption{
      Head configurations of different multi-task frameworks.
    }\label{tab:multi_task_voxel}
    \vspace*{-1em}
  \end{table}

With the goal of improving performances while reducing computational costs, recent advancements in autonomous driving research have introduced several multi-task frameworks~\cite{ye2023lidarmultinet,zhou2023lidarformer,xu2023aop}.
These methods all leverage voxel grid data representation to unify critical tasks (\ie, object detection, semantic segmentation, and panoptic segmentation) into one framework.
These tasks, however, call for various voxel types and grid resolutions, necessitating the usage of task-specific heads, as illustrated in \cref{tab:multi_task_voxel}.
Moreover, their detection heads focus on central areas of objects,  while segmentation requires predictions for all points.
In contrast, range-view representation is far less affected by task discrepancies and are therefore better suited for multi-tasks.

%% file: 3_methods.tex
\section{Methodology}\label{sec:method}


This section elaborates on the proposed framework, introducing dedicated designs for sake of efficiency and detection performance, and also highlighting its inherent versatility for seamlessly conduction of multiple tasks.

\begin{table*}[tb!]
  \centering
\renewcommand{\arraystretch}{1.05}
\renewcommand{\tabcolsep}{5pt}
  \small
  \begin{tabular}{l|c|c|c|cccc}
    \hline\toprule
    Method & Kernel Type  & \#Head   & Post-processing & Foregrounds in a Box\\
    \hline
    LaserNet~\cite{meyer2019lasernet} & Basic convolution & 1 & Mean-shift clustering + Weighted NMS & All points\\
    RangeDet~\cite{fan2021rangedet} & Meta kernel & 3 & Top-k selection + Weighted NMS & All points\\
    PPC~\cite{chai2021point} & Graph convolution & 1 & Threshold + NMS & Centroid\\
    FCOS-LiDAR~\cite{tian2022fully} & Basic convolution & 6 & Threshold + NMS + Top-k selection & Dynamic  top-k points\\
    \hline
    Ours & Basic convolution & 1 & Threshold + NMS & All points + Projected centroid \\
    \bottomrule
  \end{tabular}  
  \caption{Comparisons of different range-view-based object detectors.
  }\label{tab:method_compare_detectors}
\end{table*}


\subsection{Range View Revisited}\label{sec:method_bg}
Without loss of generality, we first recap the scanning process of single-return mechnical LiDARs.
During one sweep, a LiDAR sensor with $m$ beams equiangularly scans the environment for $n$ times, generating a range image of size $m\times n$.
Each valid pixel in the range image corresponds to a scanned point, providing multiple physical measurements, such as range value $r$, intensity $i$, and elongation $e$. 
Additionally, the direction of the scan produces the angles of azimuth $\theta$ and inclination $\phi$.
LiDAR conducts physical measurements in the spherical frame: $(r, \theta, \phi)$. The Cartesian coordinates of the 3D point can be found by $x=r\cos(\phi)\cos(\theta), y=r\cos(\phi)\sin(\theta), z=r\sin(\phi)$.

Performing perception on the range image presents two acknowledged advantages over that on the dominant voxelized representation: 
(1) The perception range is not constrained for the range-view representation, whereas the voxelized representation disregards measurements outside the pre-defined grid.
(2) The inherent compactness of range image allows for efficient feature extractions utilizing optimized operators and mature architectures designed for 2D images.
Despite these advantages, range-view-based detectors still suffer from inferior performances.
Successful attempts~\cite{fan2021rangedet,chai2021point,tian2022fully,meyer2019lasernet} (see~\cref{tab:method_compare_detectors}) narrowed the gap by giving up range view computation efficiency.

We question the necessity for such sacrifices. Our proposed SVM network bridges the detection performance gap with only a simple convolutional architecture and a single detection head design, and it naturally extends to multiple perception tasks.
For post-processing, we avoid using the inefficient top-k operator and resort to the straightforward threshold filtering scheme.
Under these constraints, we boost the \textbf{detection performance} by proposing Perspective Centric Label Assignment (PCLA) in~\cref{sec:method_ls} and View Adaptive Regression (VAR) in~\cref{sec:method_split}.
\cref{sec:method_whole} details the \textbf{scalability} of the framework by revealing its capability on multiple perception tasks.

\subsection{Perspective Centric Label Assignment}\label{sec:method_ls}

\begin{figure}[tb!]
  \centering
  \includegraphics[trim={80pt 110pt 80pt 120pt},clip, width=0.45\textwidth]{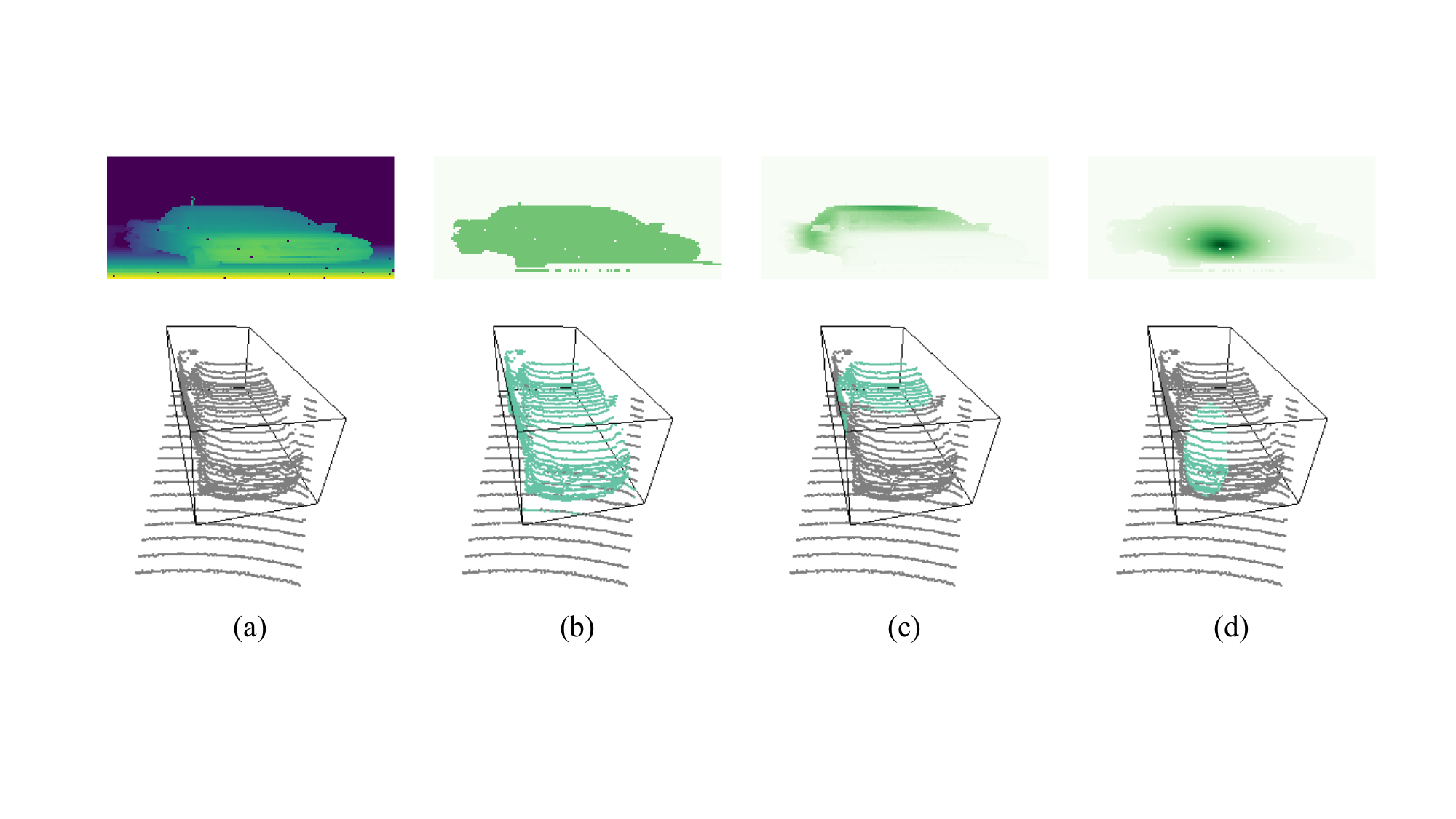}
  \caption{Comparison of different label assignments used in range-view-based detectors. 
  The top row presents the range-view results while the bottom row shows the corresponding point clouds and bounding boxes.
  (a) The raw range image and point clouds.
  (b) All points within the box are regarded as foregrounds~\cite{duan2019centernet, tian2019fcos, zhang2020bridging}.
  (c) PPC~\cite{chai2021point} assigns foregrounds with a Gaussian function, which is based on the normalized 3D distances between points and the box centers.
  (d) We propose a Perspective Centric Label Assignment which is more suitable for the range view representation.
  }
  \label{fig:method_la}
\end{figure}

Label assignment strategies can greatly impact detection performances~\cite{duan2019centernet, tian2019fcos, zhang2020bridging}.
For range-view-based object detectors~\cite{bewley2020range, meyer2019lasernet, fan2021rangedet}, a common practice is to treat all points within the 3D box as foregrounds, as shown in \cref{fig:method_la}(b).
High fidelity LiDARs can collect way too many points on the object, especially for close-by targets.
For example, in the widely adopted Waymo Open Dataset~\cite{sun2020scalability}, each frame has more than 17K points in average falling on the objects.
Naively treating all these points as foreground inevitably incurs high computation overhead in post-processing.
In addition, erroneous predictions on the edge points will further hinder the performances of object detectors.


In contrast, PPC~\cite{chai2021point} assigns classification labels based on 3D distances of points to box centers.
Specifically, given a box at $\mathbf{c}$ containing points $\mathcal{P} = \{\mathbf{p}_1, \mathbf{p}_2, \cdots, \mathbf{p}_n\}$, a point $\mathbf{p}_i$ will be assigned with the value
\begin{equation}
  \small
  s_i = \frac{\mathcal{N}(\|\mathbf{p_i} - \mathbf{c}\|_2, \sigma)}{\max_{j} \mathcal{N}(\|\mathbf{p_j} - \mathbf{c}\|_2, \sigma)}.
\end{equation}
Here, the $\sigma$ is a pre-defined parameter, and the normalization in the formula guarantees that each box has at least one positive label.
Although working well for kernels that extract features in 3D protocols, its highlighted pixels in the range image may distribute in border regions, as shown in \cref{fig:method_la}(c).
That is incompatible with the nature of 2D convolutions we intend to use, and can lead to degraded detection performances as confirmed by our experiments.

To address the defects of previous label assignment strategies, we propose the Perspective Centric Label Assignment (PCLA) as visualized in \cref{fig:method_la}(d).
The core idea behind PCLA is to assign target scores based on the projected distances of points to the corresponding box centers.
For a 3D point $\mathbf{p} = (x, y, z)$, we define the projected distance to a box center $\mathbf{c} = (x^c, y^c, z^c)$ as:
\begin{equation*}
  \small
  \begin{split}
    d(\mathbf{p}, \mathbf{c})  & = \sqrt{(\sqrt{(x- x^c)^2 + (y - y^c)^2}\cdot \cos\theta)^2 + (z-z^c)^2} \\
    & = \sqrt{(x - x^c)^2\cos^2\theta  + (y - y^c)^2\cos^2\theta + (z-z^c)^2},
  \end{split}
\end{equation*}
where $\theta=\arctan(y/x)$ is the azimuth angle for the point $\mathbf{p}$.
Our next objective is to convert the distance to a score within the range $[0, 1]$, and let short distances correspond to large score values.
To achieve this, we introduce a center-ness score in the range $[0, 1]$, akin to the FCOS~\cite{tian2019fcos} in 2D object detection.
Specifically, we first calculate projected distances of 8 box corners $\{\mathbf{cr}_i, i=1,2,\cdots, 8\}$ to the box center using $d(\mathbf{cr}_i, \mathbf{c})$.
The distance of a point $\mathbf{p}_i$, $i\in \{1, 2,\cdots, n\}$ is then normalized by
\begin{equation*}
  \small
  \hat{d}(\mathbf{p}_i, \mathbf{c}) = \max(1, \frac{d(\mathbf{p}_i, \mathbf{c}) \ }{ \max_{j\in \{0, 1, \cdots, 8\}} d(\mathbf{cr}_j, \mathbf{c})}).
\end{equation*}
In the end, the center-ness score $c_i$ is computed by 
\begin{equation*}
  \small
  c_i = \frac{1 - \hat{d}(\mathbf{p}_i, \mathbf{c})}{1 - \min_{j\in \{1, 2, \cdots, n\}}\hat{d}(\mathbf{p}_j, \mathbf{c})}.
\end{equation*}
The generated center-ness score $c_i$ distributes in the range of $[0, 1]$ as $0\leq \hat{d}(\mathbf{p}_i, \mathbf{c})\leq 1$.
In each box, the max value of 1 occurs for a point with the smallest projected distance.

Alongside the center-ness, we incorporate another design into our framework inspired by FCOS~\cite{tian2019fcos}: the simultaneous predictions of both classification and center-ness scores.
This choice stems from the understanding that, despite predicting all points within the bounding box can impact the detection performance and increase the post-processing costs, it offers valuable advantages.
One of its primary benefits lies in its ability to produce segmentation results, as demonstrated in \cref{fig:method_la}(b), owing to the intrinsic nature of point clouds. 
This capability expands the utility of the framework beyond the object detection.
Eventually, we assign a classification label $s$ for each point, and a center-ness value $c$ for a point if it belongs to an object.
The corresponding loss function is then formulated as
\begin{equation}
  \small
  L_{cls} = \sum_{i, j}(L_{1} (s_{i, j}^p, s_{i, j}) + \frac{\lambda_s}{n_{i, j}} \mathbb{I}(c_{i, j} > 0) L_{2} (c_{i, j}^p, c_{i, j})).
  \label{eq:cls}
\end{equation}
Here $\{i, j\}$ is the index of the range image and the superscript $p$ means the variable is predicted by the model.
$n_{i, j}$ is the number of points the corresponding bounding box contains.
$L_1(\cdot)$ and $L_2(\cdot)$ are two loss functions, and $\lambda_c$ is a hyper-parameter to re-weight the two losses.
$\mathbb{I}(c_{i,j}>0)$ is the indicator function, being 1 if $c_{i, j} > 0$ and 0 otherwise.

\begin{figure*}[tb!]
  \centering
  \includegraphics[trim={140pt 150pt 60pt 40pt},clip, width=0.9\textwidth]{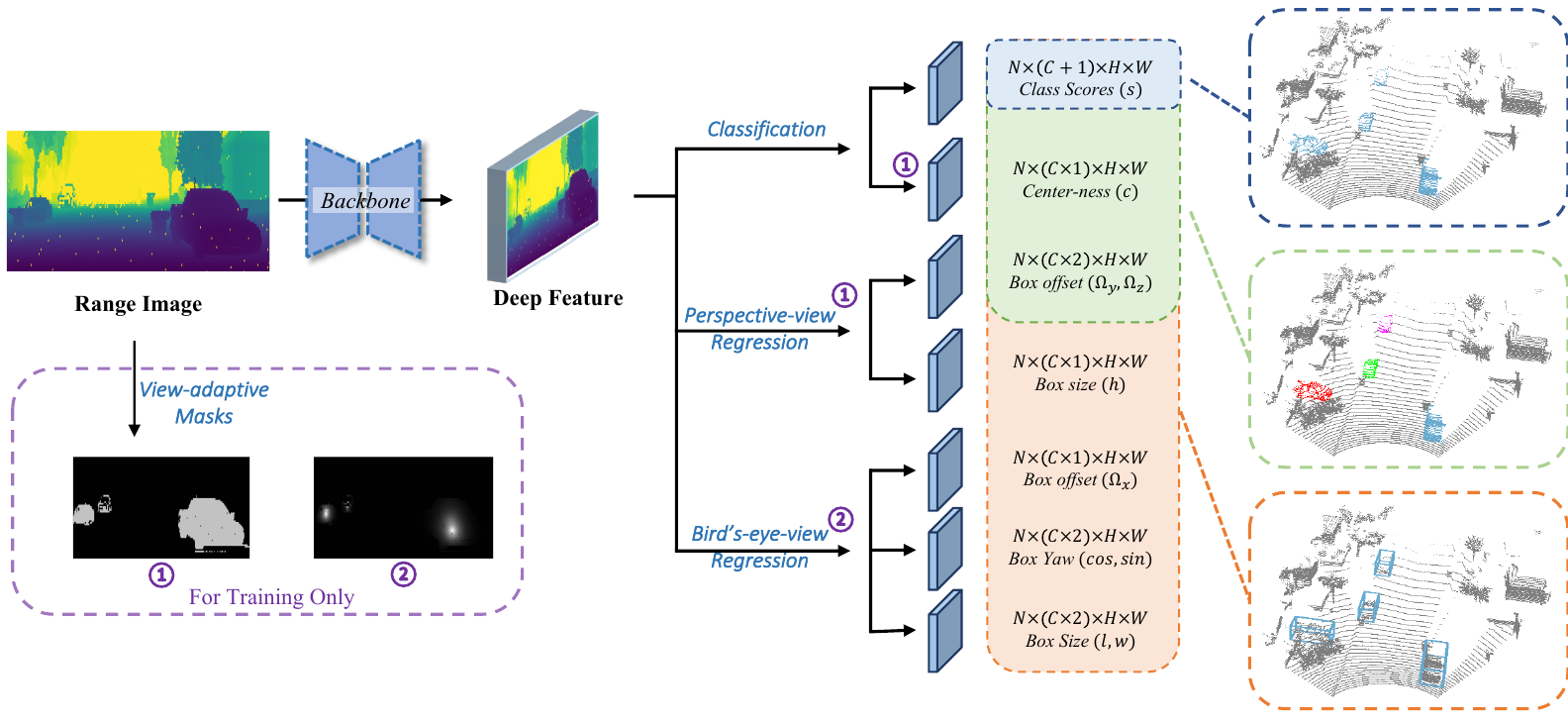}
  \caption{
  The proposed range-view-based perception system employs three branches after extracting deep features from a range image.
  The classification branch predicts semantic labels for all points, enabling the task of point cloud segmentation.
  Apart from center-ness scores from the classification branch, perspective-view regression further generates $\Omega_y$, $\Omega_z$, $h$, for valid object points.
  These results collectively contribute to the panoptic segmentation.
  In the final branch, bird's-eye-view regression is performed for the remaining elements.
  These elements are exclusively regressed for centric points within boxes, and complement the predictions of 3D detection boxes.
  }
  \label{fig:method}
\end{figure*}

\subsection{View Adaptive Regression}\label{sec:method_split}
Consider a point $\mathbf{p} = (x, y, z)$ within the bounding box $\{x^b, y^b, z^b, l^b, w^b, h^b, \theta^b\}$, where $(x^b, y^b, z^b)$ represent the box center's coordinates, $(l^b, w^b, h^b)$ denote the box extent, and $\theta^b$ is the yaw angle in the bird's-eye view. 
Following RangeDet~\cite{fan2021rangedet}, we formulate the regression targets as $\{\Omega_x, \Omega_y, \Omega_z, \log l^b, \log w^b, \log h^b, \cos \phi, \sin \phi \}$.
In this context, $\Omega_x, \Omega_y, \Omega_z, \phi$ are calculated in the view direction of the point $\mathbf{p}$ as
\begin{equation*}
  \small
  \begin{split}
    \Omega_x & = +\cos \alpha \cdot(x^b - x) + \sin \alpha\cdot (y^b - y), \\
    \Omega_y & = -\sin \alpha\cdot (x^b - x) + \cos \alpha\cdot (y^b - y), \\
    \Omega_z  & = z^b - z, \quad  \phi  = \theta^b - \alpha. \\
  \end{split}
\end{equation*}
Here $\alpha = \arctan(y/x)$ is the azimuth angle of the laser scan when measuring the point. 

\begin{figure}[t!]
  \centering
  \subfloat[Vehicle class.]{\includegraphics[width=0.24\textwidth]{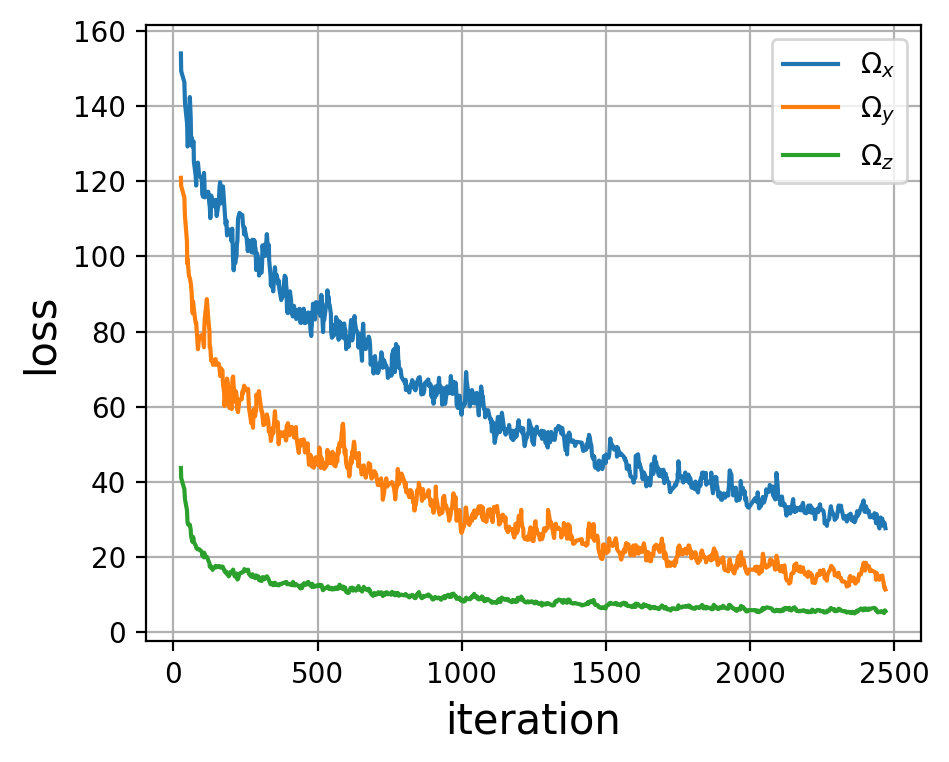}}
  \subfloat[Pedestrian class.]{\includegraphics[width=0.23\textwidth]{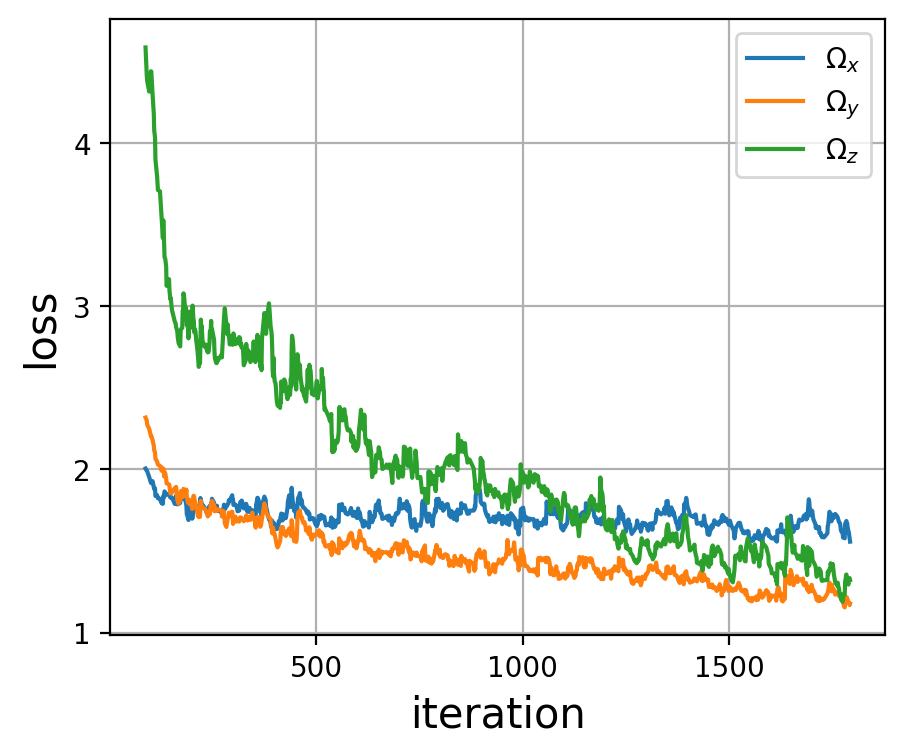}}
  \caption{
  The loss curves of $\Omega_x, \Omega_y, \Omega_z$ when training a range-view-based object detector on the vehicle and pedestrian classes in Waymo Open Dataset~\cite{sun2020scalability}.
  In both scenarios, the $\Omega_y, \Omega_z$ exhibit obviously smaller losses than $\Omega_x$ at the end of the training.
  } \label{method_loss}
\end{figure}

An interesting observation raises our attention when training the range-view-based object detectors:  $\Omega_x$ is consistently more difficult to learn than $\Omega_y$ and $\Omega_z$, as indicated by \cref{method_loss}.
Regressing on the box size also exhibits a similar pattern that $\log l^b, \log w^b$ is harder than the box height.
An intuitive explanation is that ${\Omega_y, \Omega_z, \log h^b}$ can be distinctly observed in the perspective view, whereas others are more clearly visible in the bird's-eye view.
Consequently, these regression targets may choose conflicting feature combinations during training.
This motivates us to employ two separate branches: one for regressing elements preferred in the perspective view (\ie, $\mathcal{P} = \{\Omega_y, \Omega_z, \log h^b\}$), and another for those preferred in the bird's-eye view (\ie, $\mathcal{Q} = \{\Omega_x, \log l^b, \log w^b, \cos \phi, \sin \phi \}$) respectively.

\begin{table}[tb!]
  \centering
  \renewcommand{\arraystretch}{1.05}
  \renewcommand{\tabcolsep}{5pt}
    \small
  \begin{tabular}{l|ccc}
    \toprule
      &  $\Omega_x$ & $\Omega_y$ &  $\Omega_z$ \\
      \hline
    All points & 0.117 & 0.071  &  0.042\\
    Centric points & 0.099 & 0.063  & 0.041 \\
    Edge Points & 0.132 & 0.076  & 0.043 \\
    \bottomrule
  \end{tabular}
  \caption{Mean prediction errors (meter) of $\Omega_x, \Omega_y, \Omega_z$ for all points, centric points (defined as points with center-nesses $s \geq 0.5$) and edge points (defined as $s > 0.5$) within  bounding boxes.}\label{tab:method_reg}
\end{table}

Besides the loss curves, we further analyze the prediction errors from points located differently within bounding boxes, as presented in \cref{tab:method_reg}.
Specifically, we generate predictions on the initial 1k frames from the Waymo Open Dataset~\cite{sun2020scalability} using a pre-trained object detector.
Then, we calculate the mean prediction errors of  all points, centric points and edge points from vehicle category.
Our experiments reveal that $\Omega_x$ is not only the most challenging one for regression, but also has the largest error disparity between predictions for centric and edge points. 
On the contrary, error disparities are of values $0.013$m and $0.002$m for $\Omega_y, \Omega_z$, which are much smaller than $0.033$m for $\Omega_z$.
These results indicate that regressing $\Omega_x$ for edge points is not an effective choice, motivating us to exclude them from the training.
Performing the same treatment on the $\Omega_y, \Omega_z$ should impact the detection performances marginally (also verified by later experiments), while can hinder the framework's scalability on more tasks.
Therefore, we perform a view-adaptive masking scheme on the regression loss at index $(i, j)$:
\begin{equation}
  \small
  L_{i, j}(e^p, e) = \left\{
  \begin{split}
    &\mathbb{I}_{(c_{i, j} > 0)} \cdot L_3(e^p-e), & \text{if } e\in \mathcal{P},\\
    & \mathbb{I}_{(s_{i, j} > \tau)} \cdot L_3(e^p-e), & \text{if } e\in \mathcal{Q}.\\
  \end{split}
    \right.
    \label{eq:reg_detail}
\end{equation}
Here $\tau$ is a pre-defined threshold for excluding edge points and $L_3$ is a loss function. 
Our final regression loss is
\begin{equation}
  \small
  L_{reg} = \sum_{e\in \mathcal{P}\cup \mathcal{Q}}\sum_{i, j} \frac{\lambda_r}{n_{i, j}} L_{i, j}(e^p, e).
  \label{eq:reg}
\end{equation}

\subsection{The Whole Framework}\label{sec:method_whole}
Our proposed PCLA and VAR modules place several changes to the training and inference procedures as shown in \cref{fig:method}:
(a) The classification branch predicts center-ness scores in addition to the semantic labels.
(b) The box targets are separated into two disjoint branches based on their observability.
(c) During training, the model learn elements in different locations controlled by adaptive masks.
Our designs improve the detection performances without introducing any resource-intensive modules or heavy post-processing, which ensures the efficacy of the framework.
The training loss for the entire system is $L_{all} = L_{cls} + L_{reg}$, where $L_{cls}$ and $L_{reg}$ are defined in \cref{eq:cls,eq:reg}.
We next introduce how the framework is capable of performing multiple tasks.

\noindent\textbf{Object Detection}.
During inference, we use pre-defined thresholds to filter out background points and edge points respectively, according to the semantic and center-ness predictions.
Bounding boxes are subsequently decoded from the retained points and deduplicated by NMS.
This procedure not only ensures robust detection performance but also streamlines the post-processing phase.
For example, a range image in Waymo Open Dataset~\cite{sun2020scalability} contains $~150$K valid point in average, with approximately 17K are object points.
Our method dramatically reduces the average number of points needed for decoding to just 3K, substantially easing the computational load during post-processing.

\noindent\textbf{Segmentation}.
The predicted classification labels can naturally be employed for semantic segmentation.
It's noteworthy that the number of categories for the segmentation task is typically larger than that for detection.
Simply increasing the dimension of the classification heatmap can address this issue without impacting other tasks.

\noindent\textbf{Panoptic Segmentation}. 
Our panoptic segmentation utilizes classification scores $s$, center-ness scores $c$, and the projected offsets ${\Omega_y, \Omega_z}$ for all object points. 
Due to the unique data distribution, direct clustering yields unsatisfactory results in the range-view modality. 
Therefore, we meticulously design a clustering scheme involving merging after centric grouping, as illustrated in \cref{alg:panoptic}. 
Furthermore, the absence of $\Omega_x$ causes object points to distribute along ray directions after adding offsets, rendering 3D distance unsuitable as a distance metric. 
Consequently, we propose a new distance metric tailored for the range-view modality, which essentially focuses more on distances orthogonal to the view point.
Additional details can be found in our supplementary materials.



  \begin{algorithm}[htb!]
    \KwIn{
      Points $\{p^i = (x^i, y^i, z^i)\}_1^N$, center-nesses $\{c^i\}_1^N$, offsets $\{\Omega_y^i$ and $\Omega_z^i\}_1^N$, and predicted semantic scores for the specific class $\{s^i\}_1^N$, 
    }
    \KwPara{Thresholds $\tau_c, \tau_s$.}
    \KwOut{Instance labels $\{I_i\}_1^N$.}
    For each point $p^i$, calculate the offset points by
    $\{\bar p^i = (x^i + \cos\alpha^i\cdot \Omega_y, y^i + \sin\alpha^i\cdot \Omega_y, z^i + \Omega_z^i)\}$, where $ \alpha^i = \arctan(y^i/x^i)$ is the azimuth angle \;
    Perform NMS on the center-ness heatmap and get the updated center-nesses $\{\hat c^i\}_1^N$\;
    Find the indexes of points with high center-ness values by $\mathcal{Q} = \{i \text{ if } \hat c_i > \tau_c\}$\;
    Cluster the offset points $\{\bar p^i, i \in \mathcal{Q}\}$ and get $M$ clusters $\{K_1, K_2, \cdots K_M\}$ \;
    \For{$i = 1, 2, \cdots, N$}{
      \If{$s^i\leq \tau_s$}{
        Continue\;
      }
      Calculate the distances of offset point $\bar p^i$ to each cluster $m$ by $d_{m} = \min \{D(\bar p^i, \bar p^j), j \in K_m\}$\;
      Assign the instance label for the i-th point as $I_i = \arg\min d_m $\;
    }
    \caption{Object clustering for a specific class.}\label{alg:panoptic}
  \end{algorithm}

%% file: 4_experiments.tex
\section{Experiments}\label{sec:exp}

In this section, we first detail the experimental setup in \cref{sec:exp_setup}. 
Main results and ablation studies on the 3D object detection are shown in \cref{sec:exp_main} and \cref{sec:exp_ab}, respectively.
In the end, we present some segmentation results in \cref{sec:exp_seg}.

\subsection{Experimental Setup}\label{sec:exp_setup}

\noindent\textbf{Dataset and Metrics.} 
Our experiments are conducted on large-scale Waymo Open Dataset~\cite{sun2020scalability} (WOD), which is the only dataset that provides raw range images.
The dataset contains 798 sequences for training, 202 for validation, and 150 for testing. 
Experiments in \cref{exp:main} use the entire training dataset, while others use 25\% uniformly sampled frames.
We report the 3D LEVEL 1 average precision on the WOD validation set in all experiments.

\noindent\textbf{Training Details.}
Our backbone is a lightweight convolutional network manually searched by PPC~\cite{chai2021point}. 
For the head, each branch consists of four $3\times3$ layers with 64 channels and a final prediction layer, similar to FCOS~\cite{tian2019fcos}.
Following PPC~\cite{chai2021point}, our training uses batch size 128 for 300 epochs, with an AdamW optimizer. 
One-cycle learning rate policy with an initial learning rate of 0.01 is employed. 
We adopt focal loss~\cite{lin2017focal} for $L_1$ in \cref{eq:cls}, and balanced\_L1 loss~\cite{li2021lidar} for $L_2$ in \cref{eq:cls} and $L_3$ in \cref{eq:reg_detail}.
Random global flip and ground-truth copy-paste are applied for data augmentation. 

\noindent\textbf{Other Configurations.}
Unless stated otherwise, the hyper-parameter $\tau$ in \cref{eq:cls} is set as 0.5. 
The weights $\lambda_s, \lambda_r$ in \cref{eq:cls,eq:reg} are 0.1 and 1, respectively.
The backbone from PPC~\cite{chai2021point} generates heatmaps with $2\times$ down-sample ratio.
To align with other range-view-based detectors~\cite{sun2021rsn, fan2021rangedet, meyer2019lasernet, tian2022fully}, we further add one block to up-sample the heatmap to be $1\times$ of the input resolution.
Models with the modified backbone are marked with $(1\times)$ in our results.

\subsection{Main Results}\label{sec:exp_main}
\begin{table*}[tb!]
  \centering
\renewcommand{\arraystretch}{1.05}
\renewcommand{\tabcolsep}{3pt}
  \small
  \begin{tabular}{lccccccccccc}
    \hline\toprule
    \multirow{2}{*}{Method} & \multirow{2}{*}{View} & Conv.  & \multicolumn{4}{c}{3D AP (\%) on Vehicle } &  & \multicolumn{4}{c}{3D AP (\%) on Pedestrian}  \\
        \cline{4-7}    \cline{9-12}
                           & &only    &   Overall & 0m - 30m & 30m - 50m & 50m - inf& & Overall & 0m - 30m & 30m - 50m & 50m - inf\\
    \hline
    PointPillars$^*$~\cite{lang2019pointpillars} & BEV & & 62.2& 81.8&  55.7& 31.2 && 60.0 & 68.9 & 57.6 & 46.0\\
    PointPillars$^{\P}$~\cite{lang2019pointpillars} & BEV && 71.56 &- &-&-&&70.61&-&-&- \\
    Voxel-RCNN & BEV & &75.59&  92.49 & 74.09 & 53.15 && - & - & - & -\\
    \hline
    RangeDet~\cite{fan2021rangedet} & RV & &72.85 & 87.96 & 69.03 & 48.88 && 75.94 & 82.20 & 75.39 & 65.74  \\
    PPC~\cite{chai2021point} & RV&& 65.2 &- &-&-& &73.9&-&-&-\\
    \hline
    LaserNet~\cite{meyer2019lasernet} & RV &{\textbf \checkmark}& 52.11 & 70.94 & 52.91& 29.62 && 63.4 & 73.47 &  61.55 & 42.69    \\
    RangeDet$^\dag$~\cite{fan2021rangedet}& RV &{\textbf \checkmark} &  63.57 & 84.64 &  59.54 & 38.58 && - & - & - & -\\
    PPC$^\dag$~\cite{chai2021point} & RV&{\textbf \checkmark}& 60.3 &- &-&-& &63.4&-&-&-\\
    \hline
    Ours & RV& {\textbf \checkmark}&70.31 & 85.49 & 67.74 & 48.24 && 68.23 & 72.64 & 67.59 & 58.45\\
    Ours ($1 \times$) & RV &{\textbf \checkmark} & 73.08 & 87.31 & 70.52 & 51.74 && 72.42 & 73.45 & 72.43 & 68.28 \\    
    \bottomrule
  \end{tabular}
  \caption{
    Detection performances on the validation set of the Waymo Open Dataset.
    Here, RangeDet$^\dag$~\cite{fan2021rangedet} and PPC$^\dag$~\cite{chai2021point} correspond to their models without customized kernels for feature extraction.
    *: Implemented by \cite{sun2020scalability}. $\P$: Implemented by MMDetection3D.
  }\label{exp:main}
\end{table*}

\cref{exp:main} presents performances of different LiDAR-based object detectors.
It's noteworthy that RangeDet~\cite{fan2021rangedet} and PPC~\cite{chai2021point} both introduce customized layers, which are computationally expensive and hard for deployment.
In contrast, our method achieves superior results using only vanilla convolutions.
It finally achieves the best mAP of 73.08 among all range-view-based detectors on vehicle. 

One interesting observation is that our improvements mostly occur for far objects.
For example, in the range 0m - 30m, the mAP of our model ($1\times$) is 87.31, slightly lower than 87.96 mAP from RangeDet.
However, the mAP is 51.74 for objects with distances larger than 50m, which is evidently higher than 48.8 from RangeDet.
The phenomenon is consistent for the class of pedestrians.
The mAP are 73.45 \vs 82.20 for the range 0m - 30m, while are 68.28 \vs 65.74 for the range 50m - inf.
This may be caused by the fact that RangeDet decodes boxes from points with top-k classification scores.
These points however are mostly from nearby objects, as they usually have higher scores than far objects.
In contrast, our method relieves this imbalance by disregarding edge points, thereby filtering out noisy predictions as well as involving more points from far objects.


In \cref{exp:main}, we also present performances of range-view-based object detectors without customized layers for feature aggregation (\ie, LaserNet, RangeDet$^\dag$ and PPC$^\dag$ in the table). 
When using the same backbone network, PPC$^\dag$ achieves 60.3 mAP on class vehicle and 63.4 mAP on class pedestrian, while our models have mAPs of 70.31 and 68.23.
The improvements are +10.01 and +4.87, respectively.
After updating the backbone by up-sampling the heatmaps, the final mAPs reach 73.08 and 72.42, with final improvements are +12.92 and +9.02.
Our performances also surpass RangeDet$^\dag$ by +9.51 mAP on the vehicle class, where the later model uses three heads to predict objects in different ranges, while our model utilizes a single head.
These results illustrate the effectiveness of our method.

\subsection{Ablation Studies}\label{sec:exp_ab}
This part presents ablation studies on the proposed PCLA and VAR.
All experiments here are conducted on 25\% uniformly sampled frames of the WOD training set.
The training epoch is further reduced to 36 for fast experimentations.

\begin{table}[tb!]
  \centering
\renewcommand{\arraystretch}{1.05}
\renewcommand{\tabcolsep}{3pt}
  \small
  \begin{tabular}{llc|ccc}
    \hline\toprule
    & \multirow{2}{*}{Foregrounds}  & \multicolumn{4}{c}{3D AP Vehicle (\%)}  \\
        \cline{3-6}  
       &&   \multicolumn{1}{c}{Overall} & 0m-30m & 30m-50m & 50m-inf \\
    \hline
    A & All object points  &  64.81 &	81.51 &  61.00 & 41.83 \\
    A1 & + GT Gaussian~\cite{chai2021point}
       & 64.78 & 82.23 & 60.93 & 40.94\\
    A2 & + GT Center-ness 
    & 68.68 &	83.77 & 66.24 & 47.23\\
    \hline
    B & Gaussian~\cite{chai2021point} 
    & 66.22 & 83.00 & 62.23 & 42.58 \\
    C & PCLA (Ours)  & 69.29 & 84.86 & 66.63 & 47.10\\
    \bottomrule
  \end{tabular}
  \caption{
    Results of different label assignments. GT: ground-truth.
  }\label{exp:ab_la}
\end{table}

\noindent\textbf{Effects of label assignments.}
We begin with training a model without center-ness predictions, which essentially considers all object points as foregrounds.
Besides directly benchmarking the model (A), we also remove predictions from points with low ground-truth Gaussian values (A1) and with low ground-truth center-ness values (A2).
As shown in \cref{exp:ab_la}, (A1) has a slight degradation on the overall performance compared to (A), while (A2) achieves evident improvements on the mAP, from 64.81 to 68.68.
That demonstrates that the center-ness can help filter noisy detections and therefore is suitable for the range-view modality.

Finally, we train detectors predicting Gaussian scores (B) and center-ness values (C) in an end-to-end fashion.
Model (C) outperforms (A) by 4.48 and (B) by 3.07 in mAP.  validating the efficacy of proposed PCLA.

\begin{table}[tb!]
  \centering
\renewcommand{\arraystretch}{1.05}
\renewcommand{\tabcolsep}{3pt}
  \small
  \begin{tabular}{crrcc}
    \hline\toprule
    Model & GFLOPs & \#Params & 3D AP (\%) \\
    \hline
    S1 &  5.95 & 1.77M & 68.33  \\ 
    S2 &  11.88	& 3.52M & 68.79 \\
    S3 & 17.79 & 5.28M & 69.08 \\
    \hline
    T1 &  5.85 & 1.74M & 68.72 \\
    T2 &  11.89	& 3.53M & 69.29 \\ 
    \bottomrule
  \end{tabular}
  \caption{
    Results of models with single regression branch (S1, S2, S3) and two regression branches (T1, T2).
  }\label{exp:ab_split}  
  \vspace{-1em}
\end{table}

\begin{figure*}[tb!]
  \centering  \includegraphics[width=1\textwidth]{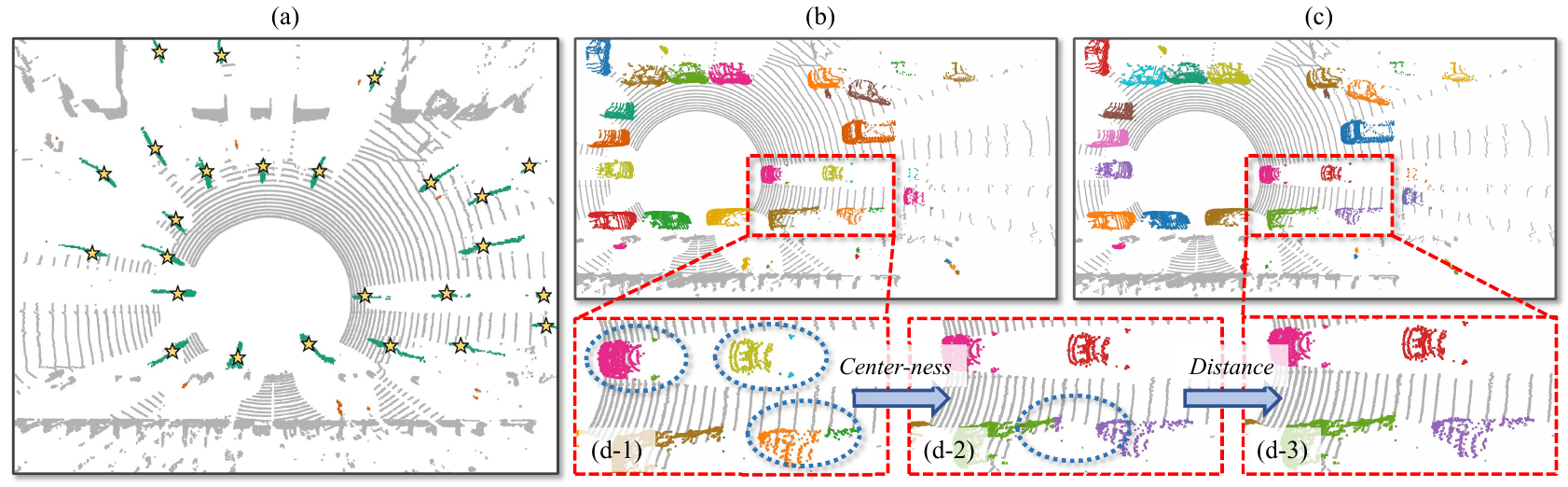}
  \caption{
    Visualization of panoptic segmentation with our framework.
    (a) Points from an object distribute in a ray direction after adding the $\Omega_y, \Omega_z$ to their 3D coordinates. 
    Stars highlight points with high center-ness scores.
    (b) Conventional clustering of offset points yields unsatisfactory results.
    (c) Performances are notably improved by the incorporation of center-ness information and a new distance metric.
  }
  \label{fig:visualization}
\end{figure*}

\noindent
\textbf{Effects of separate regression.}
The first part of VAR is assigning regression elements in two distinct branches.
To reveal its efficacy, we build models with single regression branch and two regression branches under different structures (details in the supplementary), and  report model performances, GFLOPs, number of parameters in \cref{exp:ab_split}.
With the similar GFLOPs and parameter numbers, models with two branches evidently outperform those with a single branch.
Specifically, the mAP of model (T1) is 0.39 higher than that of (S1), while the improvement of (T2) compared to its counterpart (S2) is 0.50.
Moreover, (S3) has the heaviest structures with GFLOPs nearly 50\% larger than that of (T2).
However, its AP is still lower than T2's.
All these results demonstrate the rationality of separately regressing elements based on their visibility from two viewpoints.

\begin{table}[tb!]
  \centering
\renewcommand{\arraystretch}{1.05}
\renewcommand{\tabcolsep}{3pt}
  \small
  \begin{tabular}{lcc|cccc}
    \hline\toprule
    \multirow{2}{*}{$M$ on branch(s)} & \multirow{2}{*}{$\tau$}  & \multicolumn{4}{c}{ 3D AP (\%) on Vehicle}  \\
        \cline{3-6}  
        & & Overall & 0m-30m & 30m-50m & 50m-inf\\
    \hline
    None & - & 68.86 & 84.40 & 66.01 & 46.69\\
    \hline
    PV  & 0.5 & 68.92 & 84.51 & 66.23 & 46.56\\
    \hline
     \multirow{4}{*}{BEV}  & 0.3 & 69.19 & 84.61 & \textbf{66.63} & \textbf{47.30}	\\
         & 0.5 & \textbf{69.29}	& 84.86 & \textbf{66.63} & 47.10 \\
         & 0.7 & 68.98	& 84.68 & 65.95 & 46.76\\
         & 0.9 & 68.97& \textbf{84.95} & 65.81 & 46.80\\
    \hline
    PV+BEV & 0.5 & 69.21 &84.65 & 66.43 &47.21    \\    
    \bottomrule
  \end{tabular}
  \caption{
    Model performances under different regression schemes.
  }\label{exp:ab_rc_on}    
\end{table}

\noindent
\textbf{Effects of view-adaptive masks}.
Our view-adaptive masking scheme lets $\Omega_y, \Omega_z, h$ be regressed for all object points, while letting other elements be regressed for centric points within boxes.
Denoting the mask of centric points as $M$, we report model performances when applying $M$ to different branches in \cref{exp:ab_rc_on}.
Without $M$, the overall mAP on class vehicle is 68.86.
Applying $M$ in the branch for perspective-view elements results in similar mAP to the baseline.
On the contrary, applying $M$ in the branch for bird's-eye-view elements consistently leads to performance improvements with different threshold $\tau$.
The best result is achieved when $\tau=0.5$.
When both branches use $M$, the mAP shows no big difference again.
These results demonstrate that regressing $\Omega_y, \Omega_z, h$ for all object points does not affect the performance, but provides the framework with potentials for multiple tasks.
However, the remaining elements should be regressed for centric points only for  detection performances.

\subsection{Segmentation Results}\label{sec:exp_seg}
In the main text, we present a visualization to showcase our capabilities for multiple tasks (more quantitative results in the supplementary).
In \cref{fig:visualization}(a), object points are distributed in ray directions after adding $\Omega_y, \Omega_z$ to their coordinates.
Conventional clustering methods fail to achieve satisfactory results, as depicted in \cref{fig:visualization}(b).
The first reason for this failure is that the edge points can lead to inaccurate clusters, leading to multiple clusters for a single object (\cref{fig:visualization}(d-1)).
To address this, we initially cluster the centric points and then assign the remaining foregrounds to these clusters.
Moreover, we find that some object points are still wrongly assigned to nearby objects, as shown in \cref{fig:visualization}(d-2).
This is because the 3D distance metric does not align with the unique distribution of offset points.
That motivates us to design a more suitable distance metric which pays more attention on distances orthogonal to azimuth angles.
We finally achieve satisfactory results as depicted in \cref{fig:visualization}(c).

%% file: 5_conclusions.tex
\section{Conclusions}\label{sec:conclusion}
In this work, we identify that the primary benefits of the range-view representation stem from its efficiency and the potential for multi-task applications.
To highlight these advantages, we introduce an efficient multi-tasking framework with advanced detection performances.
Equipped with the proposed Perspective Centric Label Assignment and View Adaptive Regression, our model achieves the state-of-the-art detection performances among range-view-based methods.
Several results for segmentation further reveal our method's capacities for multiple tasks.
These results verify the efficacy of our small, versatile and mighty method.

%% file: 6_supplementary.tex
\appendix
\section{Details of Panoptic Segmentation}
This section presents some details of performing panoptic segmentation using our framework.

\subsection{The Customized Distance}
\begin{figure}[htb!]
  \centering
  \includegraphics[trim={0pt 0pt 0pt 0pt},clip, width=0.4\textwidth]{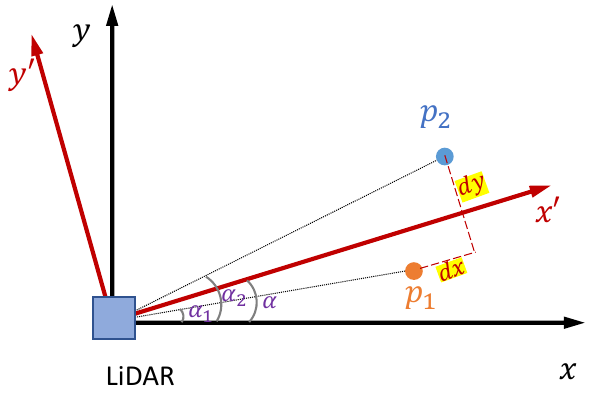}
  \caption{The proposed distance.
  }
  \label{fig:method_dist}
\end{figure}

In our work, we design a new distance metric tailored to scenarios where points are distributed along ray directions. 
In essence, our distance down-weights the distance in the view point.
Consider two points $\mathbf{p}_1 = (x_1, y_1, z_1)$ and $\mathbf{p}_2 = (x_2, y_2, z_2)$.
Instead of using the original coordinates, we introduce new coordinates for the $x$ and $y$ axes. 
Specifically, given the distinct azimuth angles for these points, we define the direction of the new $x$ axis as the mean of their azimuth angles, as illustrated in \cref{fig:method_dist}. 
Consequently, the angle for the newly-defined $x$ axis is
\begin{equation*}
  \small
  \alpha = \frac{\arctan(y_1/x_1) + \arctan(y_2/x_2)}{2}.
\end{equation*}

For a point $(x, y, z)$, its coordinates in the revised system are
\begin{equation*}
  \small
  \begin{pmatrix}
    \hat x \\ \hat y \\  \hat z
  \end{pmatrix} =
  \begin{pmatrix}
      \cos(\alpha) & \sin(\alpha) & 0 \\
      -\sin(\alpha) &  \cos(\alpha)  & 0 \\
      0                    & 0                     & 1
  \end{pmatrix} 
  \begin{pmatrix}
    x \\ y \\z
  \end{pmatrix}.
\end{equation*}
Then we can calculate the adjusted coordinates of the two points as $\mathbf{\hat p}_1 = (\hat x_1, \hat y_1, \hat z_1)$ and $\mathbf{\hat p}_2 = (\hat x_2, \hat y_2, \hat z_2)$.
Subsequently, we obtain
\begin{equation*}
  \small 
  dx = |\hat x_1 - \hat x_2 |, 
  dy = |\hat y_1 - \hat y_2 |, 
  dz = |\hat z_1 - \hat z_2 |.
\end{equation*}

Our customized distance is defined as 
\begin{equation*}
  \small
  d(\mathbf{p}_1, \mathbf{p}_2) = \sqrt{
    \lambda \cdot dx^2 + dy^2 + dz^2  
  }.
\end{equation*}
Here, we introduce the hyperparameter $\lambda$ within the range $[0, 1)$ to down-weight the impact of $dx$.

\subsection{Extension to General Datasets}
In the main text, our object center and the associated offsets rely on the annotations of bounding boxes.
However, certain object categories may lack bounding box annotations in typical scenarios.
Instead, a point can be labeled with the object class and an object identity.
In such instances, we can straightforwardly consider the average coordinates of points within an object as its object center. 
Subsequently, the offsets can be computed by evaluating the points with respect to the object center.

For the training, we only need to extend to dimension of the classification heatmap for predictions on more object classes.
The center-ness and offset regression will predict on corresponding points. 
Apart from these adjustments, the framework remains unchanged. 
This demonstrates the generality and scalability of our framework.

\section{Experiments on Object Detection}

In this section, we present several extra experiments on object detection.

\begin{table}[htb!]
  \centering
  \renewcommand{\arraystretch}{1.05}
  \renewcommand{\tabcolsep}{5pt}
    \small
    \begin{tabular}{crrcc}
      \hline\toprule
      Model & GFLOPs & \#Params & Configs. for reg. heads \\
      \hline
      S1 &  5.95 & 1.77M & [64, 64, 64, 64]\\
      S2 &  11.88	& 3.52M & [100, 100, 100, 64] \\
      S3 & 17.79 & 5.28M & [128, 128, 128, 64]\\
      \hline
      \multirow{2}{*}{T1} &  \multirow{2}{*}{5.85} & \multirow{2}{*}{1.74M} & [39, 39, 39, 64]\\
      & & & [39, 39, 39, 64] \\
      \hline
      \multirow{2}{*}{T2} &  \multirow{2}{*}{11.89}	& \multirow{2}{*}{3.53M} & [64, 64, 64, 64]\\
      & & & [64, 64, 64, 64]\\
      \bottomrule
  \end{tabular}
  \caption{
    Model details in Tab.6 of the main text. 
    Configs.: configurations. reg.: regression.
  }\label{exp:ab_hc}  
\end{table}

\noindent
\textbf{Details of model S1-3 and T1-2.}
In Tab.~6 of the main text, we present the model performances for different models.
These models have the same structures in terms of the backbone and the classification head.
Moreover, the input and output channels for regression are 64 for all models.
Models S1, S2, and S3 comprise a single regression branch, whereas T1 and T2 incorporate two branches.
The output channels of the four convolution layers in each branch can be found in \cref{exp:ab_hc}.

\begin{table}[htb!]
  \centering
\renewcommand{\arraystretch}{1.05}
\renewcommand{\tabcolsep}{4pt}
  \small
  \begin{tabular}{cc|cccc}
    \hline\toprule
     \multirow{2}{*}{Offsets}  & \multicolumn{4}{c}{3D AP (\%) on Vehicle}  \\
    \cline{2-5}
     &  Overall  & 0m-30m & 30m-50m & 50m-inf\\
    \hline
    $dx, dy, dz$ & 67.25 &	83.77 & 63.99 & 43.96 \\
    $\Omega_x, \Omega_y, \Omega_z$ &  69.29 & 84.86 & 66.63 & 47.10\\
    \bottomrule
  \end{tabular}
  \caption{
    Model performances with different offsets used.
  }\label{exp:ab_reg}  
\end{table}
\noindent

\textbf{Effects of different offsets.}
Voxel-based object detectors typically employ offsets in Euclidean space, whereas most range-view-based methods utilize offsets projected by the azimuth angle. 
Therefore, we investigate the impacts of offsets when using different coordinate spaces, as depicted in \cref{exp:ab_reg}. 
Here, $dx, dy, dz$ denote the offsets in Euclidean space, while $\Omega_x, \Omega_y, \Omega_z$ represent those used in our method. 
The latter type of offsets results in a noticeable improvement of 2.05 mAP over the former one. 
This demonstrates that $\Omega_x, \Omega_y, \Omega_z$ are more suitable for the range-view modality.

\begin{table}[htb!]
  \centering
\renewcommand{\arraystretch}{1.05}
\renewcommand{\tabcolsep}{3pt}
  \small
  \begin{tabular}{lccccccc}
    \hline\toprule
    & \multicolumn{3}{c}{Class Vehicle} && \multicolumn{3}{c}{Class Pedestrian}\\
    \cline{2-4} \cline{6-8}
    & 0-30 & 30-50 & 50-inf & & 0-30 & 30-50 & 50-inf \\
    \hline
    \#objects & 5.44 & 2.96 & 5.92 & & 0.91 & 0.86 & 1.28\\
    \#P. per obj. & 1740.6 & 145.6 & 116.6 && 131.2 & 29.1 & 24.8 \\
    \#CP. per obj. & 124.8 & 16.7 & 14.4 && 36.1 & 11.4 & 10.7 \\
    \hline
    \#CP./\#P. & 7.2\% & 11.5\% & 12.3\% && 27.8\% & 38.5\% & 43.5\%\\
    \hline 
    \bottomrule
  \end{tabular}
  \caption{
    Mean numbers of objects, points per objects and centric points per objects. P.: points. obj.: object. CP.: centric points.
  }\label{exp:main_helper}  
\end{table}

\noindent
\textbf{Distributions of objects and points.}
In the main text, we observe a notable enhancement in detection performance, particularly for distant objects, through the incorporation of our center-ness. 
Our hypothesis is that center-ness not only aids in filtering out noisy predictions but also alleviates the imbalance of objects at different distances during inference. 
To validate our hypothesis, we present several numerical results in \cref{exp:main_helper}.
Specifically, for vehicle and pedestrian classes, we report the mean numbers of objects at different distances, as well as the mean numbers of points/centric points in each object. 
We observe that the average object numbers are similar in the ranges [0m, 30m] and [50m, inf]. 
However, nearby objects contain significantly larger numbers of points than distant objects, leading to an imbalance that can overwhelm the inference with objects in close ranges.
Our center-ness mitigates this issue by filtering more points from nearby objects. For instance, 7.2\% of points are retained for vehicles in the 0m-30m range. 
In contrast, the retention ratio for vehicles with distances greater than 50m increases to 12.3\%. 
Similarly, for the pedestrian class, retention ratios are 27.8\% and 43.5\% for objects in the ranges [0m-30m] and [50m-inf], respectively. 
This highlights that our post-processing can down-weight nearby objects, leading to more significant improvements in longer distances.

\begin{table}[htb!]
  \centering
\renewcommand{\arraystretch}{1.05}
\renewcommand{\tabcolsep}{3pt}
  \small
  \begin{tabular}{cc|cccc}
    \hline\toprule
    Regression on  & \multicolumn{4}{c}{3D AP (\%) on Vehicle}  \\
    \cline{2-5}
    box centers      &  Overall  & 0m-30m & 30m-50m & 50m-inf\\
    \hline
    Absolute coordinates & 48.66 &	65.99 & 45.43 & 25.19 \\
    Relative offsets &  69.29 & 84.86 & 66.63 & 47.10\\
    \bottomrule
  \end{tabular}
  \caption{
    Model performances when directly regressing 3D coordinates of box centers and regressing  offsets to the box centers.
  }\label{exp:ab_reg}  
\end{table}

\noindent
\textbf{Effects of ways to regress box centers.}
We notice some 3D object detectors for camera images directly regress the absolute coordinates of box centers.
However, this approach is not well-suited for LiDAR data, where each point is known by its 3D coordinates. 
Leveraging this information can alleviate the learning challenges associated with the detection task.
As shown in \cref{exp:ab_reg}, directly regressing the box centers results in only a mAP of 48.66\% for the vehicle class, which is significantly lower than when using relative offsets.
Therefore, regressing the offsets is reasonable.

\section{Semantic Segmentation}
The main text describes the classification labels derived from the bounding boxes. 
Typically, there are more semantic classes for the segmentation task than those for object detection.
The framework easily accommodates this scenario by extending the dimension of the predicted classification heatmap to assign classes not present in detection labels.

To further demonstrate the capability of our used backbone and head for general semantic segmentation, we train the CENet~\cite{cheng2022cenet} in our framework on the SemanticKITTI~\cite{behley2019semantickitti} dataset. 
It's important to note that SemanticKITTI does not provide the raw range image.
We utilize this dataset solely to showcase the framework's capacity rather than to achieve optimal performances. 
Our results are presented in \cref{tab:seg}, indicating that the framework performs adequately for the task.

\begin{table*}  
  \small
  \centering
  \renewcommand{\arraystretch}{1.05}
  \renewcommand{\tabcolsep}{3pt}  
  \begin{tabular}{l|l|lllllllllllllllllll}
  \toprule
  method & mIOU & \rotatebox[origin=c]{90}{car}  
  & \rotatebox[origin=c]{90}{bicycle} 
  & \rotatebox[origin=c]{90}{motorcycle} 
  & \rotatebox[origin=c]{90}{truck} 
  & \rotatebox[origin=c]{90}{other-vehicle} 
  & \rotatebox[origin=c]{90}{person} 
  & \rotatebox[origin=c]{90}{bicycle} 
  & \rotatebox[origin=c]{90}{motorcycle} 
  & \rotatebox[origin=c]{90}{road} 
  & \rotatebox[origin=c]{90}{parking} 
  & \rotatebox[origin=c]{90}{sidewalk} 
  & \rotatebox[origin=c]{90}{other-ground} 
  & \rotatebox[origin=c]{90}{building} 
  & \rotatebox[origin=c]{90}{fence} 
  & \rotatebox[origin=c]{90}{vegetable} 
  & \rotatebox[origin=c]{90}{trunk} 
  & \rotatebox[origin=c]{90}{terrain} 
  & \rotatebox[origin=c]{90}{pole} 
  & \rotatebox[origin=c]{90}{traffic-sign}  \\
  \hline
  CENet  & 60.5 & 92.8 & 51.4    & 44.7       & 37.0  & 35.8          & 63.5   & 59.7    & 28.0       & 90.9 & 62.5    & 74.5     & 26.8         & 88.2     & 59.8  & 81.6      & 66.8  & 65.2    & 56.6 & 63.1         \\
  \hline
  CENet$^*$   & 57.8 & 92.0 & 44.5    & 35.9       & 36.7  & 29.6          & 56.6   & 49.0    & 20.3       & 91.6 & 64.0    & 75.5     & 25.8         & 89.4     & 60.7  & 81.5      & 64.5  & 65.3    & 54.8 & 59.6         \\
  \bottomrule
  \end{tabular}
  \caption{
    Performances of the original CENet and that in our framework (denoted with CENet$^*$).
  }\label{tab:seg}
\end{table*}

\section{Panoptic Segmentation}
Panoptic segmentation integrates semantic segmentation on stuff classes and instance segmentation on thing classes. 
Given that our framework readily supports semantic segmentation, we only need to demonstrate its feasibility by focusing on instance segmentation for the thing objects. 
Consequently, we select the vehicle class in the Waymo Open Dataset~\cite{sun2020scalability}, as other datasets for panoptic segmentation do not provide the raw range image.

\noindent\textbf{Evaluation Metric.}
Following current panoptic segmentation methods~\cite{milioto2020lidar, sirohi2021efficientlps}, we use the  Mean Panoptic Quality (PQ) as the primary metric, which can be decomposed into Segmentation Quality (SQ) and Recognition Quality (RQ) to provide additional insights of the results.
The formulation of PQ is 
\begin{equation}
  \small 
  PQ = \underbrace{
    \frac{\sum_{(p, g)\in TP_c IoU(p, g)}}{|TP|}
    }_{\text{Segmentation Quality (SQ)}} 
  \times \underbrace{
    \frac{|TP|}{|TP| + \frac{1}{2}|FP| + \frac{1}{2}|FN|}
  }_{\text{Recognition Quality (RQ)}}.
\end{equation}
Here, $(p, g)$ represent the prediction and ground truth, and $TP, FP, FN$ are the set of true positive, false positives, and false negative matches. 
A match is a true positive if the corresponding IoU is larger than 0.5.

\begin{table}[htb!]
  \centering
  \renewcommand{\arraystretch}{1.05}
  \renewcommand{\tabcolsep}{5pt}
    \small
  \begin{tabular}{l|ccc}
    \toprule
      Method & PQ & SQ & RQ \\
      \hline
      Bounding box & 81.25 & 84.90 & 95.70 \\
      \hline
      Clustering on $\Omega_y, \Omega_z$ & 77.26 & 82.31 & 93.86 \\
      Ours$^*$ & 81.86 & 87.50 & 93.55 \\
      Ours & 83.72 & 88.28 & 94.84 \\
    \bottomrule
  \end{tabular}
  \caption{
    Performances of different methods on the vehicle class. 
    Here Ours$^*$ and Ours represent our clustering methods with 3D distance and customized distance, respectively.
  }\label{tab:exp_ps}
\end{table}

\noindent
\textbf{Comparison of different schemes.}
Using the predicted results from our model (denoted as \(x_1\) in our main results), we conduct instance segmentation on the vehicle class with bounding boxes, clustering using \(\Omega_y, \Omega_z\), our algorithm with 3D distance, and our algorithm with a customized distance.
It is evident that directly clustering offset points with \(\Omega_y, \Omega_z\) does not yield satisfactory results. 
The PQ value is largely behind that of using bounding boxes. 
However, with the introduction of center-ness and our merge-after-clustering scheme, the PQ value increases to 81.86, surpassing the baseline value of 81.25.
It's important to note that performing instance segmentation with bounding boxes is computationally intensive, as it requires point-in-box checking for all points and all boxes. 
In contrast, we only perform clustering on the centric points, while the merging steps are highly efficient. 
With our customized distance metric, the PQ value increases to 83.72, representing a significant improvement of 2.53 compared to the baseline.

\begin{table}[htb!]
  \centering
  \renewcommand{\arraystretch}{1.05}
  \renewcommand{\tabcolsep}{6pt}
    \small
  \begin{tabular}{l|ccc}
    \toprule
      $\lambda$ & PQ & SQ & RQ \\
      \hline
      0 & 74.64 & 81.07 & 92.07 \\
      1e-5\qquad & 78.30 & 84.26 & 92.93 \\
      1e-4 & 81.58 & 86.86 & 93.92 \\
      1e-3 & 83.18 & 87.90 & 94.63 \\
      1e-2 & 83.72 & 88.28 & 94.84 \\
      1e-1 & 83.36 & 88.09 & 94.93 \\
      1 &  81.86 & 87.50 & 93.55 \\
    \bottomrule
  \end{tabular}
  \caption{
    The effects of the hyperparameter $\lambda$.
  }\label{tab:exp_ps_ablation}
\end{table}

\noindent
\textbf{Ablation study on hyperparameter $\lambda$.}
Our ablation study on our distance metric is shown in \cref{tab:exp_ps_ablation}.
It is evident that using no information along the view direction results in the poorest PQ value of 74.64.
The PQ value initially increases and then decreases with an increase in \(\lambda\).
The best result is achieved when \(\lambda=1e-2\).

%% file: arxiv_svm.bbl
\begin{thebibliography}{45}
\providecommand{\natexlab}[1]{#1}
\providecommand{\url}[1]{\texttt{#1}}
\expandafter\ifx\csname urlstyle\endcsname\relax
  \providecommand{\doi}[1]{doi: #1}\else
  \providecommand{\doi}{doi: \begingroup \urlstyle{rm}\Url}\fi

\bibitem[Ando et~al.(2023)Ando, Gidaris, Bursuc, Puy, Boulch, and
  Marlet]{ando2023rangevit}
Angelika Ando, Spyros Gidaris, Andrei Bursuc, Gilles Puy, Alexandre Boulch, and
  Renaud Marlet.
\newblock Rangevit: Towards vision transformers for 3d semantic segmentation in
  autonomous driving.
\newblock In \emph{IEEE Conference on Computer Vision and Pattern Recognition},
  pages 5240--5250, 2023.

\bibitem[Behley et~al.(2019)Behley, Garbade, Milioto, Quenzel, Behnke,
  Stachniss, and Gall]{behley2019semantickitti}
Jens Behley, Martin Garbade, Andres Milioto, Jan Quenzel, Sven Behnke, Cyrill
  Stachniss, and Jurgen Gall.
\newblock Semantickitti: A dataset for semantic scene understanding of lidar
  sequences.
\newblock In \emph{International Conference on Computer Vision}, pages
  9297--9307, 2019.

\bibitem[Bewley et~al.(2020)Bewley, Sun, Mensink, Anguelov, and
  Sminchisescu]{bewley2020range}
Alex Bewley, Pei Sun, Thomas Mensink, Dragomir Anguelov, and Cristian
  Sminchisescu.
\newblock Range conditioned dilated convolutions for scale invariant 3d object
  detection.
\newblock \emph{arXiv preprint arXiv:2005.09927}, 2020.

\bibitem[Chai et~al.(2021)Chai, Sun, Ngiam, Wang, Caine, Vasudevan, Zhang, and
  Anguelov]{chai2021point}
Yuning Chai, Pei Sun, Jiquan Ngiam, Weiyue Wang, Benjamin Caine, Vijay
  Vasudevan, Xiao Zhang, and Dragomir Anguelov.
\newblock To the point: Efficient 3d object detection in the range image with
  graph convolution kernels.
\newblock In \emph{IEEE Conference on Computer Vision and Pattern Recognition},
  pages 16000--16009, 2021.

\bibitem[Chen et~al.(2019)Chen, Liu, Shen, and Jia]{chen2019fast}
Yilun Chen, Shu Liu, Xiaoyong Shen, and Jiaya Jia.
\newblock Fast point r-cnn.
\newblock In \emph{International Conference on Computer Vision}, pages
  9775--9784, 2019.

\bibitem[Cheng et~al.(2022)Cheng, Han, and Xiao]{cheng2022cenet}
Hui-Xian Cheng, Xian-Feng Han, and Guo-Qiang Xiao.
\newblock Cenet: Toward concise and efficient lidar semantic segmentation for
  autonomous driving.
\newblock In \emph{2022 IEEE International Conference on Multimedia and Expo
  (ICME)}, pages 01--06. IEEE, 2022.

\bibitem[Cortinhal et~al.(2020)Cortinhal, Tzelepis, and
  Erdal~Aksoy]{cortinhal2020salsanext}
Tiago Cortinhal, George Tzelepis, and Eren Erdal~Aksoy.
\newblock Salsanext: Fast, uncertainty-aware semantic segmentation of lidar
  point clouds.
\newblock In \emph{Advances in Visual Computing: 15th International Symposium,
  ISVC 2020, San Diego, CA, USA, October 5--7, 2020, Proceedings, Part II 15},
  pages 207--222. Springer, 2020.

\bibitem[Deng et~al.(2021)Deng, Shi, Li, Zhou, Zhang, and Li]{deng2021voxel}
Jiajun Deng, Shaoshuai Shi, Peiwei Li, Wengang Zhou, Yanyong Zhang, and
  Houqiang Li.
\newblock Voxel r-cnn: Towards high performance voxel-based 3d object
  detection.
\newblock In \emph{AAAI Conference on Artificial Intelligence}, pages
  1201--1209, 2021.

\bibitem[Dosovitskiy et~al.(2020)Dosovitskiy, Beyer, Kolesnikov, Weissenborn,
  Zhai, Unterthiner, Dehghani, Minderer, Heigold, Gelly,
  et~al.]{dosovitskiy2020image}
Alexey Dosovitskiy, Lucas Beyer, Alexander Kolesnikov, Dirk Weissenborn,
  Xiaohua Zhai, Thomas Unterthiner, Mostafa Dehghani, Matthias Minderer, Georg
  Heigold, Sylvain Gelly, et~al.
\newblock An image is worth 16x16 words: Transformers for image recognition at
  scale.
\newblock \emph{arXiv preprint arXiv:2010.11929}, 2020.

\bibitem[Duan et~al.(2019)Duan, Bai, Xie, Qi, Huang, and
  Tian]{duan2019centernet}
Kaiwen Duan, Song Bai, Lingxi Xie, Honggang Qi, Qingming Huang, and Qi Tian.
\newblock Centernet: Keypoint triplets for object detection.
\newblock In \emph{International Conference on Computer Vision}, pages
  6569--6578, 2019.

\bibitem[Fan et~al.(2021)Fan, Xiong, Wang, Wang, and Zhang]{fan2021rangedet}
Lue Fan, Xuan Xiong, Feng Wang, Naiyan Wang, and Zhaoxiang Zhang.
\newblock Rangedet: In defense of range view for lidar-based 3d object
  detection.
\newblock In \emph{International Conference on Computer Vision}, pages
  2918--2927, 2021.

\bibitem[Fan et~al.(2022)Fan, Pang, Zhang, Wang, Zhao, Wang, Wang, and
  Zhang]{fan2022embracing}
Lue Fan, Ziqi Pang, Tianyuan Zhang, Yu-Xiong Wang, Hang Zhao, Feng Wang, Naiyan
  Wang, and Zhaoxiang Zhang.
\newblock Embracing single stride 3d object detector with sparse transformer.
\newblock In \emph{IEEE Conference on Computer Vision and Pattern Recognition},
  pages 8458--8468, 2022.

\bibitem[Kochanov et~al.(2020)Kochanov, Nejadasl, and
  Booij]{kochanov2020kprnet}
Deyvid Kochanov, Fatemeh~Karimi Nejadasl, and Olaf Booij.
\newblock Kprnet: Improving projection-based lidar semantic segmentation.
\newblock \emph{arXiv preprint arXiv:2007.12668}, 2020.

\bibitem[Kong et~al.(2023)Kong, Liu, Chen, Ma, Zhu, Li, Hou, Qiao, and
  Liu]{kong2023rethinking}
Lingdong Kong, Youquan Liu, Runnan Chen, Yuexin Ma, Xinge Zhu, Yikang Li,
  Yuenan Hou, Yu Qiao, and Ziwei Liu.
\newblock Rethinking range view representation for lidar segmentation.
\newblock In \emph{International Conference on Computer Vision}, pages
  228--240, 2023.

\bibitem[Lang et~al.(2019)Lang, Vora, Caesar, Zhou, Yang, and
  Beijbom]{lang2019pointpillars}
Alex~H Lang, Sourabh Vora, Holger Caesar, Lubing Zhou, Jiong Yang, and Oscar
  Beijbom.
\newblock Pointpillars: Fast encoders for object detection from point clouds.
\newblock In \emph{IEEE Conference on Computer Vision and Pattern Recognition},
  pages 12697--12705, 2019.

\bibitem[Li et~al.(2023)Li, Luo, and Yang]{li2023pillarnext}
Jinyu Li, Chenxu Luo, and Xiaodong Yang.
\newblock Pillarnext: Rethinking network designs for 3d object detection in
  lidar point clouds.
\newblock In \emph{IEEE Conference on Computer Vision and Pattern Recognition},
  pages 17567--17576, 2023.

\bibitem[Li et~al.(2021)Li, Wang, and Wang]{li2021lidar}
Zhichao Li, Feng Wang, and Naiyan Wang.
\newblock Lidar r-cnn: An efficient and universal 3d object detector.
\newblock In \emph{IEEE Conference on Computer Vision and Pattern Recognition},
  pages 7546--7555, 2021.

\bibitem[Lin et~al.(2017)Lin, Goyal, Girshick, He, and
  Doll{\'a}r]{lin2017focal}
Tsung-Yi Lin, Priya Goyal, Ross Girshick, Kaiming He, and Piotr Doll{\'a}r.
\newblock Focal loss for dense object detection.
\newblock In \emph{International Conference on Computer Vision}, pages
  2980--2988, 2017.

\bibitem[Meyer et~al.(2019)Meyer, Laddha, Kee, Vallespi-Gonzalez, and
  Wellington]{meyer2019lasernet}
Gregory~P Meyer, Ankit Laddha, Eric Kee, Carlos Vallespi-Gonzalez, and Carl~K
  Wellington.
\newblock Lasernet: An efficient probabilistic 3d object detector for
  autonomous driving.
\newblock In \emph{IEEE Conference on Computer Vision and Pattern Recognition},
  pages 12677--12686, 2019.

\bibitem[Milioto et~al.(2019)Milioto, Vizzo, Behley, and
  Stachniss]{milioto2019rangenet++}
Andres Milioto, Ignacio Vizzo, Jens Behley, and Cyrill Stachniss.
\newblock Rangenet++: Fast and accurate lidar semantic segmentation.
\newblock In \emph{IEEE/RSJ international conference on intelligent robots and
  systems (IROS)}, pages 4213--4220. IEEE, 2019.

\bibitem[Milioto et~al.(2020)Milioto, Behley, McCool, and
  Stachniss]{milioto2020lidar}
Andres Milioto, Jens Behley, Chris McCool, and Cyrill Stachniss.
\newblock Lidar panoptic segmentation for autonomous driving.
\newblock In \emph{IEEE/RSJ International Conference on Intelligent Robots and
  Systems (IROS)}, pages 8505--8512. IEEE, 2020.

\bibitem[Qi et~al.(2018)Qi, Liu, Wu, Su, and Guibas]{qi2018frustum}
Charles~R Qi, Wei Liu, Chenxia Wu, Hao Su, and Leonidas~J Guibas.
\newblock Frustum pointnets for 3d object detection from rgb-d data.
\newblock In \emph{IEEE Conference on Computer Vision and Pattern Recognition},
  pages 918--927, 2018.

\bibitem[Qi et~al.(2019)Qi, Litany, He, and Guibas]{qi2019deep}
Charles~R Qi, Or Litany, Kaiming He, and Leonidas~J Guibas.
\newblock Deep hough voting for 3d object detection in point clouds.
\newblock In \emph{International Conference on Computer Vision}, pages
  9277--9286, 2019.

\bibitem[Sautier et~al.(2023)Sautier, Puy, Boulch, Marlet, and
  Lepetit]{sautier2023bevcontrast}
Corentin Sautier, Gilles Puy, Alexandre Boulch, Renaud Marlet, and Vincent
  Lepetit.
\newblock Bevcontrast: Self-supervision in bev space for automotive lidar point
  clouds.
\newblock \emph{arXiv preprint arXiv:2310.17281}, 2023.

\bibitem[Shi et~al.(2019{\natexlab{a}})Shi, Wang, and Li]{shi2019pointrcnn}
Shaoshuai Shi, Xiaogang Wang, and Hongsheng Li.
\newblock Pointrcnn: 3d object proposal generation and detection from point
  cloud.
\newblock In \emph{IEEE Conference on Computer Vision and Pattern Recognition},
  pages 770--779, 2019{\natexlab{a}}.

\bibitem[Shi et~al.(2019{\natexlab{b}})Shi, Wang, Wang, and Li]{shi2019part}
Shaoshuai Shi, Zhe Wang, Xiaogang Wang, and Hongsheng Li.
\newblock Part-aˆ 2 net: 3d part-aware and aggregation neural network for
  object detection from point cloud.
\newblock \emph{arXiv preprint arXiv:1907.03670}, 2\penalty0 (3),
  2019{\natexlab{b}}.

\bibitem[Shi and Rajkumar(2020)]{shi2020point}
Weijing Shi and Raj Rajkumar.
\newblock Point-gnn: Graph neural network for 3d object detection in a point
  cloud.
\newblock In \emph{IEEE Conference on Computer Vision and Pattern Recognition},
  pages 1711--1719, 2020.

\bibitem[Sirohi et~al.(2021)Sirohi, Mohan, B{\"u}scher, Burgard, and
  Valada]{sirohi2021efficientlps}
Kshitij Sirohi, Rohit Mohan, Daniel B{\"u}scher, Wolfram Burgard, and Abhinav
  Valada.
\newblock Efficientlps: Efficient lidar panoptic segmentation.
\newblock \emph{IEEE Transactions on Robotics}, 38\penalty0 (3):\penalty0
  1894--1914, 2021.

\bibitem[Sun et~al.(2020)Sun, Kretzschmar, Dotiwalla, Chouard, Patnaik, Tsui,
  Guo, Zhou, Chai, Caine, et~al.]{sun2020scalability}
Pei Sun, Henrik Kretzschmar, Xerxes Dotiwalla, Aurelien Chouard, Vijaysai
  Patnaik, Paul Tsui, James Guo, Yin Zhou, Yuning Chai, Benjamin Caine, et~al.
\newblock Scalability in perception for autonomous driving: Waymo open dataset.
\newblock In \emph{IEEE Conference on Computer Vision and Pattern Recognition},
  pages 2446--2454, 2020.

\bibitem[Sun et~al.(2021)Sun, Wang, Chai, Elsayed, Bewley, Zhang, Sminchisescu,
  and Anguelov]{sun2021rsn}
Pei Sun, Weiyue Wang, Yuning Chai, Gamaleldin Elsayed, Alex Bewley, Xiao Zhang,
  Cristian Sminchisescu, and Dragomir Anguelov.
\newblock Rsn: Range sparse net for efficient, accurate lidar 3d object
  detection.
\newblock In \emph{IEEE Conference on Computer Vision and Pattern Recognition},
  pages 5725--5734, 2021.

\bibitem[Tian et~al.(2019)Tian, Shen, Chen, and He]{tian2019fcos}
Zhi Tian, Chunhua Shen, Hao Chen, and Tong He.
\newblock Fcos: Fully convolutional one-stage object detection.
\newblock In \emph{International Conference on Computer Vision}, pages
  9627--9636, 2019.

\bibitem[Tian et~al.(2022)Tian, Chu, Wang, Wei, and Shen]{tian2022fully}
Zhi Tian, Xiangxiang Chu, Xiaoming Wang, Xiaolin Wei, and Chunhua Shen.
\newblock Fully convolutional one-stage 3d object detection on lidar range
  images.
\newblock \emph{Annual Conference on Neural Information Processing Systems},
  35:\penalty0 34899--34911, 2022.

\bibitem[Xu et~al.(2020)Xu, Wu, Wang, Zhan, Vajda, Keutzer, and
  Tomizuka]{xu2020squeezesegv3}
Chenfeng Xu, Bichen Wu, Zining Wang, Wei Zhan, Peter Vajda, Kurt Keutzer, and
  Masayoshi Tomizuka.
\newblock Squeezesegv3: Spatially-adaptive convolution for efficient
  point-cloud segmentation.
\newblock In \emph{European Conference on Computer Vision}, pages 1--19.
  Springer, 2020.

\bibitem[Xu et~al.(2023)Xu, Fazlali, Ren, and Liu]{xu2023aop}
Yixuan Xu, Hamidreza Fazlali, Yuan Ren, and Bingbing Liu.
\newblock Aop-net: All-in-one perception network for lidar-based joint 3d
  object detection and panoptic segmentation.
\newblock In \emph{IEEE Intelligent Vehicles Symposium (IV)}, pages 1--7. IEEE,
  2023.

\bibitem[Yan et~al.(2018)Yan, Mao, and Li]{yan2018second}
Yan Yan, Yuxing Mao, and Bo Li.
\newblock Second: Sparsely embedded convolutional detection.
\newblock \emph{Sensors}, 18\penalty0 (10):\penalty0 3337, 2018.

\bibitem[Yang et~al.(2018)Yang, Sun, Liu, Shen, and Jia]{yang2018ipod}
Zetong Yang, Yanan Sun, Shu Liu, Xiaoyong Shen, and Jiaya Jia.
\newblock Ipod: Intensive point-based object detector for point cloud.
\newblock \emph{arXiv preprint arXiv:1812.05276}, 2018.

\bibitem[Yang et~al.(2020)Yang, Sun, Liu, and Jia]{yang20203dssd}
Zetong Yang, Yanan Sun, Shu Liu, and Jiaya Jia.
\newblock 3dssd: Point-based 3d single stage object detector.
\newblock In \emph{IEEE Conference on Computer Vision and Pattern Recognition},
  pages 11040--11048, 2020.

\bibitem[Ye et~al.(2023)Ye, Zhou, Chen, Xie, Wang, Wang, and
  Foroosh]{ye2023lidarmultinet}
Dongqiangzi Ye, Zixiang Zhou, Weijia Chen, Yufei Xie, Yu Wang, Panqu Wang, and
  Hassan Foroosh.
\newblock Lidarmultinet: Towards a unified multi-task network for lidar
  perception.
\newblock In \emph{AAAI Conference on Artificial Intelligence}, pages
  3231--3240, 2023.

\bibitem[Yin et~al.(2021)Yin, Zhou, and Krahenbuhl]{yin2021center}
Tianwei Yin, Xingyi Zhou, and Philipp Krahenbuhl.
\newblock Center-based 3d object detection and tracking.
\newblock In \emph{IEEE Conference on Computer Vision and Pattern Recognition},
  pages 11784--11793, 2021.

\bibitem[Zhang et~al.(2020{\natexlab{a}})Zhang, Chi, Yao, Lei, and
  Li]{zhang2020bridging}
Shifeng Zhang, Cheng Chi, Yongqiang Yao, Zhen Lei, and Stan~Z Li.
\newblock Bridging the gap between anchor-based and anchor-free detection via
  adaptive training sample selection.
\newblock In \emph{IEEE Conference on Computer Vision and Pattern Recognition},
  pages 9759--9768, 2020{\natexlab{a}}.

\bibitem[Zhang et~al.(2020{\natexlab{b}})Zhang, Zhou, David, Yue, Xi, Gong, and
  Foroosh]{zhang2020polarnet}
Yang Zhang, Zixiang Zhou, Philip David, Xiangyu Yue, Zerong Xi, Boqing Gong,
  and Hassan Foroosh.
\newblock Polarnet: An improved grid representation for online lidar point
  clouds semantic segmentation.
\newblock In \emph{IEEE Conference on Computer Vision and Pattern Recognition},
  pages 9601--9610, 2020{\natexlab{b}}.

\bibitem[Zhou and Tuzel(2018)]{zhou2018voxelnet}
Yin Zhou and Oncel Tuzel.
\newblock Voxelnet: End-to-end learning for point cloud based 3d object
  detection.
\newblock In \emph{IEEE Conference on Computer Vision and Pattern Recognition},
  pages 4490--4499, 2018.

\bibitem[Zhou et~al.(2022)Zhou, Zhao, Wang, Wang, and
  Foroosh]{zhou2022centerformer}
Zixiang Zhou, Xiangchen Zhao, Yu Wang, Panqu Wang, and Hassan Foroosh.
\newblock Centerformer: Center-based transformer for 3d object detection.
\newblock In \emph{European Conference on Computer Vision}, pages 496--513.
  Springer, 2022.

\bibitem[Zhou et~al.(2023)Zhou, Ye, Chen, Xie, Wang, Wang, and
  Foroosh]{zhou2023lidarformer}
Zixiang Zhou, Dongqiangzi Ye, Weijia Chen, Yufei Xie, Yu Wang, Panqu Wang, and
  Hassan Foroosh.
\newblock Lidarformer: A unified transformer-based multi-task network for lidar
  perception.
\newblock \emph{arXiv preprint arXiv:2303.12194}, 2023.

\bibitem[Zhu et~al.(2023)Zhu, Meng, Wang, Wang, Yan, and
  Yang]{zhu2023curricular}
Ziyue Zhu, Qiang Meng, Xiao Wang, Ke Wang, Liujiang Yan, and Jian Yang.
\newblock Curricular object manipulation in lidar-based object detection.
\newblock In \emph{IEEE Conference on Computer Vision and Pattern Recognition},
  pages 1125--1135, 2023.

\end{thebibliography}
